%% file: main.tex
\documentclass{article}

\usepackage{iclr2027_conference,times}

\usepackage[utf8]{inputenc}
\usepackage[T1]{fontenc}
\usepackage{microtype}
\usepackage{graphicx}
\usepackage{booktabs}
\usepackage{array}
\usepackage{multirow}
\usepackage{float}
\usepackage{subcaption}
\usepackage{amsmath}
\usepackage{amssymb}
\usepackage{mathtools}
\usepackage{amsthm}
\usepackage{url}
\usepackage{xcolor}
\usepackage{wrapfig}
\usepackage{hyperref}
\hypersetup{hidelinks}
\usepackage{tikz}
\usetikzlibrary{arrows.meta,calc,fit,positioning}

\input{vpack}

\input{math_commands}

\newsavebox{\flarecosttable}

\title{FLARE++: Low-rank attention with attention-synthesized routing}

\author{%
  Vedant Puri,
  Sri Datta Ganesh Bandreddi,
  Yongjie Jessica Zhang
  \& Levent Burak Kara \\
  Department of Mechanical Engineering \\
  Carnegie Mellon University \\
  Pittsburgh, Pennsylvania, USA
}

\iclrfinalcopy

\begin{document}

\maketitle
\ificlrfinal\lhead{Preprint.}\fi

\begin{abstract}
Full self-attention is a strong token mixer for PDE surrogates on irregular domains, but its quadratic cost limits its use on large problems.
Latent-space attention methods such as PerceiverIO, Transolver, and FLARE (Fast Low-rank Attention Routing Engine) avoid that cost by routing attention among $N$ tokens through $M\ll N$ learned latents.
They compress by dot-product matching of the input tokens against $M$ learned query tokens or projection weights: once trained, the same learned templates serve every input.
We remove this restriction with FLARE++, a low-rank attention architecture with input-conditioned routing queries.
FLARE++ uses FLARE's own encoder to map the $N$ input tokens to $M$ query tokens, which correct the learned queries.
The adapted queries then determine how that same input is compressed and redistributed.
This preserves FLARE's explicit low-rank factorization and linear $\mathcal O(NM)$ complexity, and expresses the complete routing operation with standard scaled dot-product attention (SDPA) calls alone.
We also provide a multi-GPU context-parallel implementation that shards input tokens across devices without ever gathering the full token sequence on one of them.
FLARE++ reduces FLARE's error by $25\%$ on average across five standard PDE benchmarks, achieving the lowest errors among the efficient models compared.
The gains persist on industrial-scale DrivAerML aerodynamics and on Long Range Arena, where average accuracy rises by $3.5\%$ over fixed-query FLARE.
Code is available at \url{https://github.com/vpuri3/FLARE.py}.
\end{abstract}

\section{Introduction}
\begin{figure}[!t]
\centering
\includegraphics[width=\linewidth]{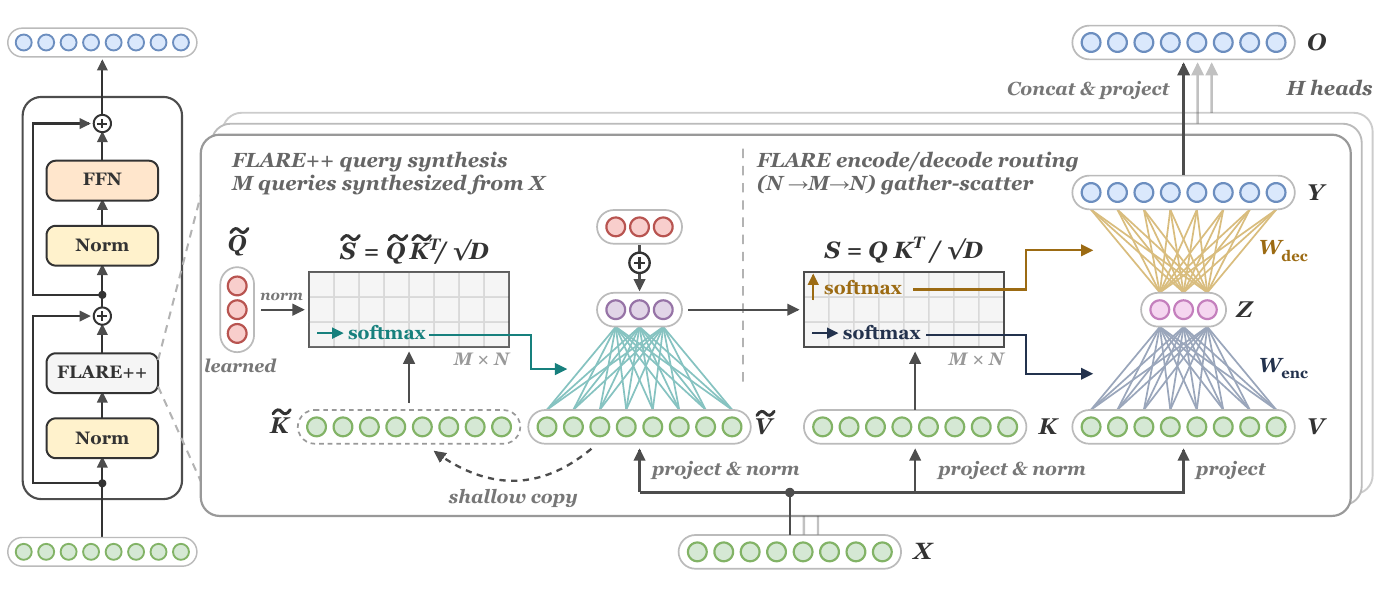}

\vspace{-0.0em}

\sbox{\flarecosttable}{\footnotesize\setlength{\tabcolsep}{3pt}%
\begin{tabular}{@{}lll@{}}
\toprule
\textbf{Operation} & \textbf{Time Complexity} & \textbf{Space Complexity} \\
\midrule
\texttt{K,V,\~{K} = project(X)} & $\mathcal{O}(3NC^2)$ & $\mathcal{O}(3NC)$ \\
\texttt{Q = Qf + SDPA(\~{Q},\~{K},\~{K})} & $\mathcal{O}(NMC)$ & $\mathcal{O}(MC)$ \\
\texttt{Z = SDPA(Q, K, V)} & $\mathcal{O}(NMC)$ & $\mathcal{O}(MC)$ \\
\texttt{Y = SDPA(K, Q, Z)} & $\mathcal{O}(NMC)$ & $\mathcal{O}(NC)$ \\
\texttt{O = merge(Y) @ Wo} & $\mathcal{O}(NC^2)$ & $\mathcal{O}(NC)$ \\
\midrule
FLARE++ layer & $\mathcal{O}(N(4C^2\!+\!3MC))$ & $\mathcal{O}(5NC\!+\!2MC)$ \\
\bottomrule
\end{tabular}}%
\noindent\hbox to \linewidth{%
  \usebox{\flarecosttable}\hfill
  \raisebox{-\dp\flarecosttable}[\ht\flarecosttable][\dp\flarecosttable]{%
    \includegraphics[height=\dimexpr\ht\flarecosttable+\dp\flarecosttable\relax,
      trim={10.74bp 7.11bp 10.85bp 5.48bp}, clip]{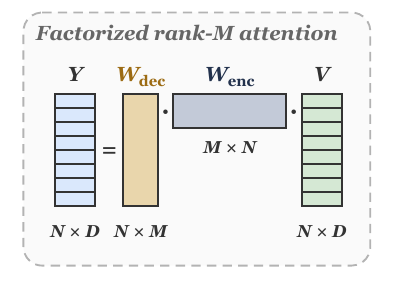}}%
}

\caption{
The FLARE++ mixer.
Learned seeds $\widetilde Q_h$ attend to the input to synthesize $M$ corrections, gated onto learned queries $Q^{\mathrm f}_h$ to form the routing queries $Q_h(X)$ (left), which gather the $N$ tokens into $M$ latent values and redistribute them (right).
The synthesis call shares its keys and values.
Arrows mark the softmax axis of each score matrix: $W_{\mathrm{enc},h}$ over the $N$ tokens, $W_{\mathrm{dec},h}$ over the $M$ routes.
Below: the mixer over all $H$ heads with its cost, and the rank-$M$ operator its two routing calls compose to.
\texttt{SDPA} is fused, so $S_h$ and $\widetilde S_h$ are never materialized.
}
\label{fig:flarepp_block}
\end{figure}

PDE surrogates must connect local physics to global field behavior.
Self-attention provides this global communication directly: each of the $N$ discretization points can attend to every other.
Its quadratic cost, however, makes this approach impractical on high-resolution engineering meshes~\citep{vaswani2017attention}.
A range of efficient token mixers communicate instead through $M\ll N$ latent tokens.
This makes global interaction affordable, but introduces a bottleneck in global communication that determines what information to collect and disseminate.

Latent-space attention models, including Perceiver IO~\citep{jaegle2021perceiverio}, the Transolver family~\citep{wu2024transolver,luotransolver++,zhou2026transolver3}, and Latent Neural Operator (LNO)~\citep{wang2024latent}, gather input features into a latent subsequence, process information in that space, and project it back to the full input domain.
FLARE (Fast Low-rank Attention Routing Engine)~\citep{puri2026flare}, which is another latent attention architecture, instead uses latent space to effect low-rank self-attention.
Its $M\ll N$ learned queries gather $N$ point tokens into $M$ latent tokens with a cross-attention call, and a second cross-attention call redistributes those tokens to the points, without an intervening latent self-attention stage.
The two calls form an encode--decode operator of rank at most $M$ per head, implemented entirely with standard fused scaled dot-product attention (SDPA).
By isolating compression and decompression as routing operations, FLARE is a natural starting point.

These latent-space attention architectures compress the input field using dot-product scores against learned weights or latent tokens; in FLARE, these are the $M$ latent query tokens.
We call the addressing vectors that define this compression a \emph{compression template}.
These methods differ in how they construct this template: FLARE's queries are learned parameters, while other constructions use a learned pointwise map or a prescribed field-dependent rule.
In FLARE, training learns $M$ query templates per layer and head, and those same templates then serve every geometry and boundary condition.
The routing weights respond to the current keys, but the queries defining the template do not see the field they compress.
This fixed template determines how the rank-$M$ bottleneck is spent.

We present \textbf{FLARE++}, which constructs the compression template from the input tokens instead of fixing it after training.
This dynamic query construction reuses FLARE's own encode mechanism: learned latent seeds act as the queries of one extra encode call, gathering the $N$ input tokens into $M$ tokens.
These tokens define the routing queries of the encode--decode pair that compresses and redistributes the field (\autoref{fig:flarepp_block}).
Because query construction uses the same encode operation as FLARE, it inherits the efficient fused SDPA implementation and requires no new kernel.
One attention call constructs the routes; two more gather and dispatch values along them.

The change is confined to the mixer, and within the mixer to the construction of the compression template.
The induced routing matrix remains rank at most $M$ per head for each fixed input, and the complexity remains $\mathcal O(NM)$ for fixed channel width.
The residual backbone retains the same number of blocks, channel width, and residual update structure, so the comparison with FLARE isolates the routing change without making the network deeper or wider.
The mixer itself gains one projection and one attention call, costing $1.1$--$1.2\times$ FLARE's time per full training run (\autoref{sec:drivaerml_scaling}).

We also make this global routing practical on meshes that exceed one accelerator's memory.
Sharding tokens across GPUs needs just the standard log-sum-exp merge of blockwise and Ring Attention~\citep{liu2023blockwise,liu2024ringattention}, exchanging latent outputs and softmax statistics.
FLARE++ invokes it for query synthesis and again for value gathering, while dispatch remains local.
This supports full-surface DrivAerML evaluation on $8.2$M cells.

Our experiments compare FLARE++ with PDE-specific and general-purpose attention models.
Across five standard PDE benchmarks, FLARE++ reduces FLARE's error by $25\%$ on average and achieves the lowest errors among the efficient models compared.
Depth and latent-budget sweeps show that the gain persists as either resource grows, and that at a fixed latent budget, synthesized routing queries reach a given accuracy with a shallower model.
On full-surface DrivAerML, FLARE++ improves on FLARE in all $18$ matched depth and latent-budget configurations, by up to $33\%$ in shallow models, and Long Range Arena shows the gains extend beyond physical fields.
Our contributions are:

\begin{itemize}
\item
\textbf{Low-rank attention with synthesized routing queries.}
FLARE++ uses one extra attention call to construct routing queries from the current input, then uses them to gather and redistribute its tokens, yielding efficient rank-$\leq M$ attention with linear cost in token count.
\item
\textbf{Exact context parallelism for large meshes.}
Because tokens interact only through $M$ latents, FLARE++ shards tokens across GPUs and exchanges only $\mathcal{O}(M)$ latent outputs and softmax statistics.
This enables single-pass inference on the $8.2$M-cell DrivAerML surface.
\item
\textbf{Improved accuracy.}
FLARE++ improves on FLARE across benchmarks and is the strongest efficient attention model compared; ablations show synthesized routing queries reach a given accuracy with a shallower model than fixed ones.
\end{itemize}

\section{Related work}
\label{sec:related_work}

\begin{figure}[t]
\centering
\includegraphics[width=1.0\linewidth]{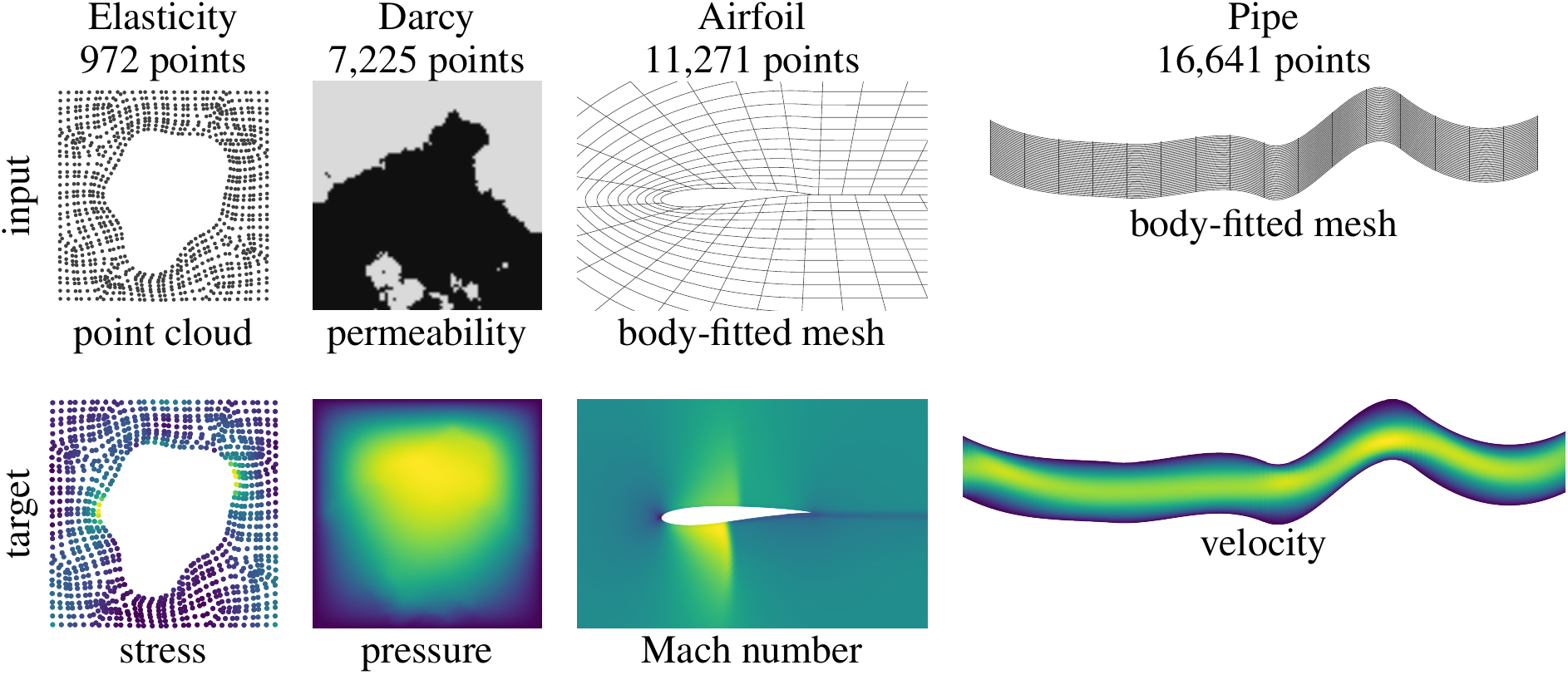}
\caption{
The two-dimensional benchmarks.
Top: the input each surrogate receives, namely a point cloud on Elasticity, the permeability field on Darcy, and body-fitted meshes on Airfoil and Pipe.
Bottom: the target field.
One case each at true aspect ratio; Airfoil is cropped from its far-field mesh.
}
\label{fig:gallery_2d}
\end{figure}

\paragraph{Neural operators on irregular domains.}
Neural operators learn maps between function spaces and extend from regular grids to irregular point sets and complex geometries~\citep{li2020fourier,lu2021learning,kovachki2023neural,pfaff2020learning,mousavi2025rigno}.
Closest to ours are transformer operators on unstructured meshes, such as GNOT~\citep{hao2023gnot}, and the geometry encoders GINO and GINOT~\citep{li2023geometry,liu2025ginot}.
We study the token mixer inside such an operator, and classify latent mixers by where their compression template comes from.

\paragraph{Learned templates.}
Perceiver IO~\citep{jaegle2021perceiverio}, LNO~\citep{wang2024latent}, Set Transformer~\citep{lee2019settransformer}, the Transolver family~\citep{wu2024transolver,luotransolver++,zhou2026transolver3}, FLARE~\citep{puri2026flare}, Luna~\citep{ma2021luna}, and TokenLearner~\citep{ryoo2021tokenlearner} compress against a learned template that is fixed at inference.
In each, latent $m$ keeps its identity across inputs.
Perceiver, LNO, and Transolver additionally apply nonlinear operations in the latent space.

\paragraph{Data-derived templates.}
Other methods use the data itself as the template, selected by a prescribed rule.
Nystr\"omformer uses segment means of the queries and keys as landmarks~\citep{xiong2021nystromformer}, Agent Attention pools the query tokens into agent tokens~\citep{han2024agent}, UPT aggregates the input onto supernodes sampled from the input points~\citep{alkin2024universal}, and AB-UPT attends through sampled anchor points~\citep{alkin2025abuptscalingneuralcfd}; hashed and clustered attention group tokens by similarity~\citep{kitaev2020reformer,vyas2020clustered}.
These templates change with every input, but their latents carry no identity across inputs: anchor $m$ on one geometry bears no relation to anchor $m$ on another, and segment means and pooling rely on token order or grids absent on unstructured meshes.

\paragraph{Recurrent latent state.}
In Slot Attention~\citep{locatello2020slot}, the original Perceiver's iterated cross-attention~\citep{jaegle2021perceiver}, Q-Former~\citep{li2023blip2}, and Luna~\citep{ma2021luna}, the latent content produced by one cross-attention becomes the next one's query.
FLARE++ carries no latent state across blocks: each block rebuilds its routing queries from a learned template, and the path that constructs them is separate from the path that gathers values.

\paragraph{Other efficient attention.}
Kernel feature maps linearize attention without a latent bottleneck~\citep{katharopoulos2020transformers,choromanski2020rethinking,zhang2024hedgehog}; \autoref{app:lra} compares several such mixers.
Unlike hypernetwork and FiLM conditioning, which generate weights or feature modulations~\citep{perez2018film,cho_hypernetwork-based_2023}, FLARE++ conditions only the attention queries, as a bounded deformation of a learned basis.

Existing latent mixers thus use either a learned template fixed at inference or a data-derived template without persistent latent identities.
FLARE++ combines the two: it keeps a learned template and adapts it to each input.
Because this adaptation is built by attention rather than by sampling or pooling, it is permutation-invariant and defined for any discretization.

\section{Method}
\label{sec:method}

\begin{table}[!t]
\centering
\caption{
Test relative $L^2$ error (in \%) on standard PDE benchmarks; bold and underline mark best and second-best among efficient models.
Full self-attention is shown separately as a quadratic-cost reference, and $\sim$ marks runs prohibitively slow.
$^{*}$Mean over ten random seeds (\autoref{app:seed_variance}).
}
\label{tab:pde_benchmark_summary}
\setlength{\tabcolsep}{0pt}
\begin{tabular*}{\textwidth}{@{\extracolsep{\fill}}lccccc@{}}
\toprule
\textbf{Model} &
\shortstack{\textbf{Elasticity}\\(1K points)} &
\shortstack{\textbf{Darcy}\\(7K points)} &
\shortstack{\textbf{Airfoil}\\(11K points)} &
\shortstack{\textbf{Pipe}\\(16K points)} &
\shortstack{\textbf{DrivAerML-40K}\\(40K points)} \\
\midrule
Full self-attention &
0.41 & 0.43 & 0.58 & $\sim$ & $\sim$ \\
\midrule
PerceiverIO &
2.80 & 2.06 & 0.77 & 0.69 & 24.80 \\
Set Transformer &
\underline{0.57} & 1.18 & 0.67 & 0.39 & \underline{6.56} \\
GNOT &
1.33 & 1.69 & 10.30 & 0.59 & 11.50 \\
LNO &
0.93 & 0.76 & 1.78 & 0.81 & 14.60 \\
Luna &
0.59 & 0.84 & 0.60 & \underline{0.38} & 6.90 \\
Transolver &
0.72 & 1.04 & 0.60 & \underline{0.38} & 7.39 \\
Transolver++ &
1.02 & 2.76 & 1.27 & 0.76 & 9.39 \\
Transolver-3 &
0.68 & 1.03 & 0.73 & 0.44 & 7.36 \\
AB-UPT &
1.34 & 2.70 & 2.93 & 4.12 & 10.0 \\
FLARE &
0.71$^{*}$ & \underline{0.76} & \underline{0.57} & 0.51 & 7.22 \\
\textbf{FLARE++ (ours)} &
\textbf{0.39}$^{*}$ & \textbf{0.62} & \textbf{0.50} & \textbf{0.35} & \textbf{5.80} \\
\bottomrule
\end{tabular*}
\end{table}

\subsection{Preliminaries}

We consider PDE surrogates that operate on fields discretized at $N$ spatial points, each carrying problem-dependent inputs such as coordinates, boundary conditions, material parameters, or an initial state.
A pointwise input projection embeds these into $N$ tokens, $X=[x_1,\ldots,x_N]^\top\in\mathbb{R}^{N\times C}$, with hidden width $C$.
Residual token-mixing and feedforward (FFN) blocks communicate information between tokens, and a pointwise output projection decodes the solution field.
We review full self-attention and FLARE's fixed-query operator, treating heads independently as in \citet{puri2026flare}.

\paragraph{Full self-attention.}
Learned projections construct $Q=XW_Q$, $K=XW_K$, $V=XW_V$, split into $H$ heads of width $D=C/H$, and head $h$ returns $Y_h=\operatorname{softmax}(Q_hK_h^\top/\sqrt D)V_h$; the head outputs are concatenated and projected by $W_O$.
Both sides of the $N\times N$ score matrix depend on the input, and each token can attend directly to every other token, at $\mathcal{O}(N^2D)$ work per head.
We write $\operatorname{SDPA}$ for this primitive as a function of its three arguments,
\begin{equation}
\label{eq:sdpa_definition}
    \operatorname{SDPA}(Q,K,V)
    =\operatorname{softmax}\!\left(\frac{QK^\top}{\sqrt D}\right)V
    \in\mathbb{R}^{N_q\times D},
    \qquad
    Q\in\mathbb{R}^{N_q\times D},\quad
    K,V\in\mathbb{R}^{N_k\times D},
\end{equation}
where the softmax normalizes over the $N_k$ keys.
Self-attention is the case $N_q=N_k=N$; the operators below take $N_q\neq N_k$ to move information between $N$ tokens and $M$ latent routes.

\paragraph{FLARE: fixed-query latent routing.}
FLARE~\citep{puri2026flare} replaces the dense matrix with an encode--decode factorization through $M\ll N$ latent routes.
Only keys and values are projected from the input; each head owns an independent learned query set:
\begin{equation}
    K_h=XW_{K,h}\in\mathbb{R}^{N\times D},\qquad
    V_h=XW_{V,h}\in\mathbb{R}^{N\times D},\qquad
    Q_h\in\mathbb{R}^{M\times D}\ \text{is learned}.
\end{equation}
The encoder and decoder are standard SDPA calls,
\begin{equation}
\label{eq:prelim_flare_sdpa}
    Z_h=\operatorname{SDPA}(Q_h,K_h,V_h),\qquad
    Y_h=\operatorname{SDPA}(K_h,Q_h,Z_h),
\end{equation}
which, writing $S_h=Q_hK_h^\top/\sqrt D\in\mathbb{R}^{M\times N}$, $W_{\mathrm{enc},h}=\operatorname{softmax}(S_h)$ and $W_{\mathrm{dec},h}=\operatorname{softmax}(S_h^\top)$, gather the $N$ input values into $M$ latent values and scatter them back:
\begin{equation}
\label{eq:prelim_flare_operator}
    Y_h=W_{\mathrm{eff},h}V_h,
    \qquad
    W_{\mathrm{eff},h}=W_{\mathrm{dec},h}W_{\mathrm{enc},h}\in\mathbb{R}^{N\times N}.
\end{equation}
Because $W_{\mathrm{eff},h}$ factors through $M$ latent routes, its rank is at most $M$, and it is never materialized: the two SDPA calls apply its factors sequentially at $\mathcal{O}(NMD)$ cost, but the queries $Q_h$ that define those routes are fixed across samples.

\subsection{FLARE++: routing-query synthesis}
\label{sec:flarepp}

\begin{figure}[!t]
\centering
\includegraphics[width=\linewidth]{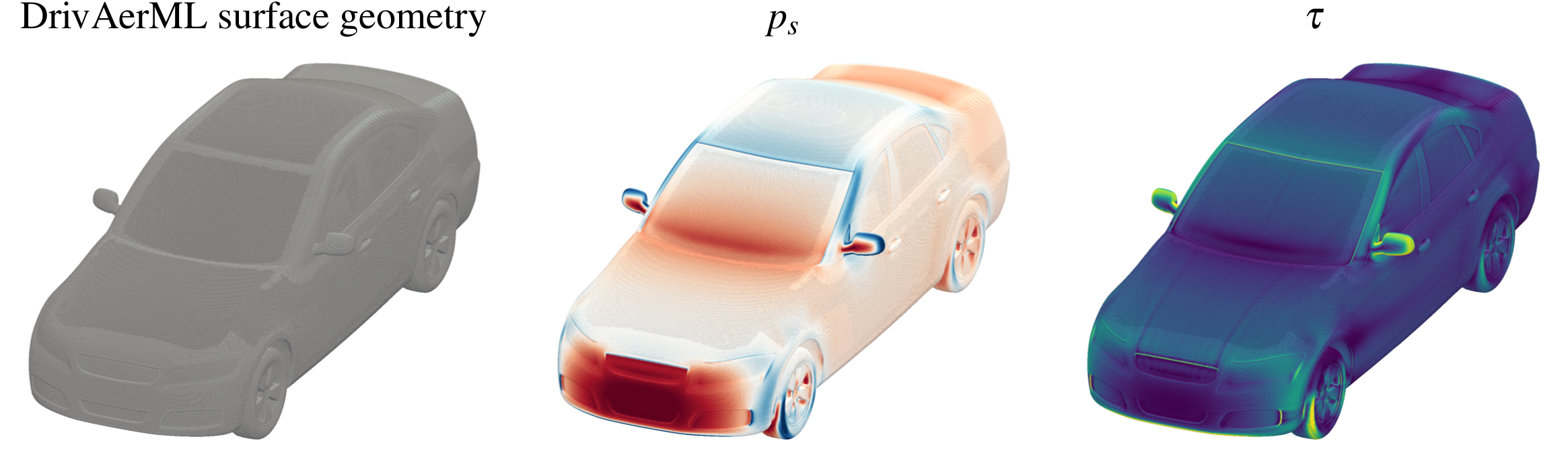}
\caption{
The DrivAerML benchmark: surface geometry, surface pressure $p_s$, and wall-shear magnitude $\tau$ on the full surface mesh of $8.2$M cells (colour scales in \autoref{app:qualitative}).
This is the full-surface task of \autoref{sec:drivaerml_scaling}, distinct from the $40$K-point DrivAerML-40K of \autoref{tab:pde_benchmark_summary}.
}
\label{fig:gallery_drivaerml}
\end{figure}

The queries $Q_h$ define how the $M$ latent routes score the input tokens.
We retain the $M$-route bottleneck and its learned queries, and let the current input adapt them.
Because the output of the synthesis call becomes the query of the routing calls, the routing vectors move while the encoder is still learning to use them, which makes early training unstable (\autoref{app:flarepp_difficulty}).
We control this by normalizing every tensor that enters an attention score per head with RMSNorm~\citep{zhang2019root,henry2020qknorm,dehghani2023vit22b}, and by anchoring the synthesized queries to a fixed learned reference, as described next (\autoref{app:flarepp_normalized}, \autoref{app:flarepp_blended}).
We write $\mathcal N_a$ for RMSNorm with a learned scale and $\mathcal N_0$ for RMSNorm without one, so that every row of $\mathcal N_0(\cdot)$ has unit root-mean-square (RMS) norm.
To construct the adaptation, we apply FLARE's encoder a second time, with learned seeds $\widetilde Q_h\in\mathbb{R}^{M\times D}$ as its queries and normalized synthesis keys, shared with the values,
\begin{equation}
\label{eq:flarepp_query_synthesis}
    \widetilde K_h=\mathcal N_0(X\widetilde W_{K,h}),
    \qquad \widetilde V_h=\widetilde K_h.
\end{equation}
Sharing them costs no substantial accuracy (\autoref{app:flarepp_shared_kv}).
The encoder's $M$ outputs are sample-specific query corrections,
\begin{align}
\label{eq:flarepp_dynamic_query}
    Q^{\mathrm d}_h(X)
    =\operatorname{SDPA}\big(\mathcal N_a(\widetilde Q_h),\widetilde K_h,\widetilde V_h\big)
    =\operatorname{softmax}\!\left(
        \frac{\mathcal N_a(\widetilde Q_h)\,\widetilde K_h^\top}{\sqrt D}
      \right)\widetilde V_h
      \in\mathbb{R}^{M\times D}.
\end{align}
Each row of $Q^{\mathrm d}_h(X)$ is a convex combination of unit-RMS value rows, so the triangle inequality gives $\operatorname{RMS}(Q^{\mathrm d}_{h,m})\le1$.
The correction is thus bounded without being normalized, and its magnitude records how coherently the attended features agree (\autoref{app:flarepp_blended}).
The routing queries combine a learned reference set $Q^{\mathrm f}_h\in\mathbb{R}^{M\times D}$, as in FLARE, with this correction through a learned per-head gate $g_h=\sigma(\gamma_h)\in(0,1)$:
\begin{equation}
\label{eq:flarepp_anchored_query}
    Q_h(X)=\mathcal N_0(Q^{\mathrm f}_h)+g_h\,Q^{\mathrm d}_h(X).
\end{equation}
The fixed term has unit RMS and provides a common routing basis; the synthesized term, of RMS at most $g_h<1$, adapts it to the current field, so every routing query has RMS between $1-g_h$ and $1+g_h$.
As $g_h\to0$, the construction approaches fixed-query routing.
Routing with $Q^{\mathrm d}_h(X)$ alone gives essentially the same FP32 accuracy as the gated form, reducing FLARE's error by $24\%$ without the routing-key and fixed-query normalizations, so those normalizations do not explain the gain; adding them to FLARE did not improve it (\autoref{app:simplified_flarepp}, \autoref{app:flarepp_blended}).
The gate matters in the mixed-precision study of \autoref{app:precision}: FLARE++ holds its FP32 accuracy in FP16 and trains without difficulty, whereas the ungated variant degrades and becomes hard to train (\autoref{app:flarepp_difficulty}, \autoref{app:flarepp_blended}).
Each synthesized correction aggregates features across the current field, so changing one part of the field can change the queries that address every token in the block.
With routing keys $K_h=\mathcal N_a(XW_{K,h})$, whose learned scale sets the temperature of the routing logits, and values $V_h=XW_{V,h}$, the routing queries enter both factors of the FLARE operator:
\begin{equation}
\label{eq:flarepp_mixer}
    Z_h=\operatorname{SDPA}(Q_h(X),K_h,V_h),\qquad
    Y_h=\operatorname{SDPA}(K_h,Q_h(X),Z_h).
\end{equation}
Writing the resulting routing factors as $W_{\mathrm{enc},h}(X)$ and $W_{\mathrm{dec},h}(X)$ gives
\begin{equation}
    Y_h=W_{\mathrm{dec},h}W_{\mathrm{enc},h}V_h,
    \qquad
    \operatorname{rank}(W_{\mathrm{dec},h}W_{\mathrm{enc},h})\leq M,
\end{equation}
so the current field determines the queries used for both gathering and dispatch.
The bound holds for each fixed input; the map remains nonlinear because the routing weights depend on $X$.
The full mixer is three fused SDPA calls (\autoref{fig:flarepp_block}), with no custom kernel or sequence-projection matrix.
Every added component (synthesis projection, seeds, gate, per-head normalizations) acts only through the routing scores $Q_hK_h^\top$; the value, output, and feed-forward paths are FLARE's (\autoref{tab:mixer_ops}).

\paragraph{Computational complexity.}

Both mixers are built from two primitives: a pointwise projection, at $\mathcal{O}(NC^2)$ time and $\mathcal{O}(NC)$ space, and a token--latent SDPA call, at $\mathcal{O}(NMC)$ time and $\mathcal{O}(NC)$ space, since a fused kernel never materializes the $N\times M$ scores.
FLARE uses three projections ($K$, $V$, output) and two SDPA calls, $\mathcal{O}(N(3C^2+2MC))$ per block; FLARE++ adds $\widetilde K$ and one call, $\mathcal{O}(N(4C^2+3MC))$, and the gated sum adds $\mathcal{O}(MC)$.
For fixed $M$ and $C$, both are linear in $N$, against $\mathcal{O}(N^2C)$ for full self-attention.
The additional projection and attention call increase mixer arithmetic by a factor between $4/3$ and $3/2$, depending on $M/C$.
\autoref{sec:experiments_efficiency} measures the corresponding time and memory cost and relates it to the accuracy gained at a given depth.

\subsection{Multi-GPU context parallelism}
\label{sec:multigpu}

Linear complexity alone does not make a high-resolution mesh fit on one accelerator, because pointwise activations still grow with $N$.
We therefore shard the token dimension across $R$ ranks, rank $r$ holding $X_r\in\mathbb{R}^{N_r\times C}$ with $\sum_r N_r=N$, which leaves projections, normalization, FFN layers, and residual updates local.
The only globally coupled primitive is the encoder gathering sharded tokens into $M$ replicated latents.
Each rank runs a fused SDPA call on its local keys and values, returning a local output $Z_r$ and rowwise log-normalizer $L_r$; the exact global result is the standard log-sum-exp merge of blockwise attention~\citep{liu2023blockwise,liu2024ringattention,dao2023flashdecoding},
\begin{equation}
\label{eq:distributed_flare_encoder}
    L=\log\sum_{r=0}^{R-1}\exp(L_r),
    \qquad
    Z=\sum_{r=0}^{R-1}\exp(L_r-L)\,Z_r,
\end{equation}
which equals the encoder applied to the concatenated sequence in exact arithmetic.
With $M$ replicated latent queries, the merge is one all-reduce of latent outputs and softmax statistics, $\mathcal{O}(HM(D+2))$ values per rank independent of $N$; reduction order can change rounding.
Decoding, $Y_r=\operatorname{SDPA}(K_r,Q,Z)$, is local once $Q$ and $Z$ are replicated, so no all-gather over tokens is ever required.
FLARE invokes the encoder once per mixer; FLARE++ invokes it twice, for $Q(X)$ and for $Z$.
The backward pass follows the same decomposition into local attention and global reductions (\autoref{app:multigpu}); \autoref{sec:experiments_efficiency} evaluates the scaling.

\section{Experiments}
\label{sec:experiments}

\begin{figure}[!t]
\centering
\includegraphics[width=\linewidth]{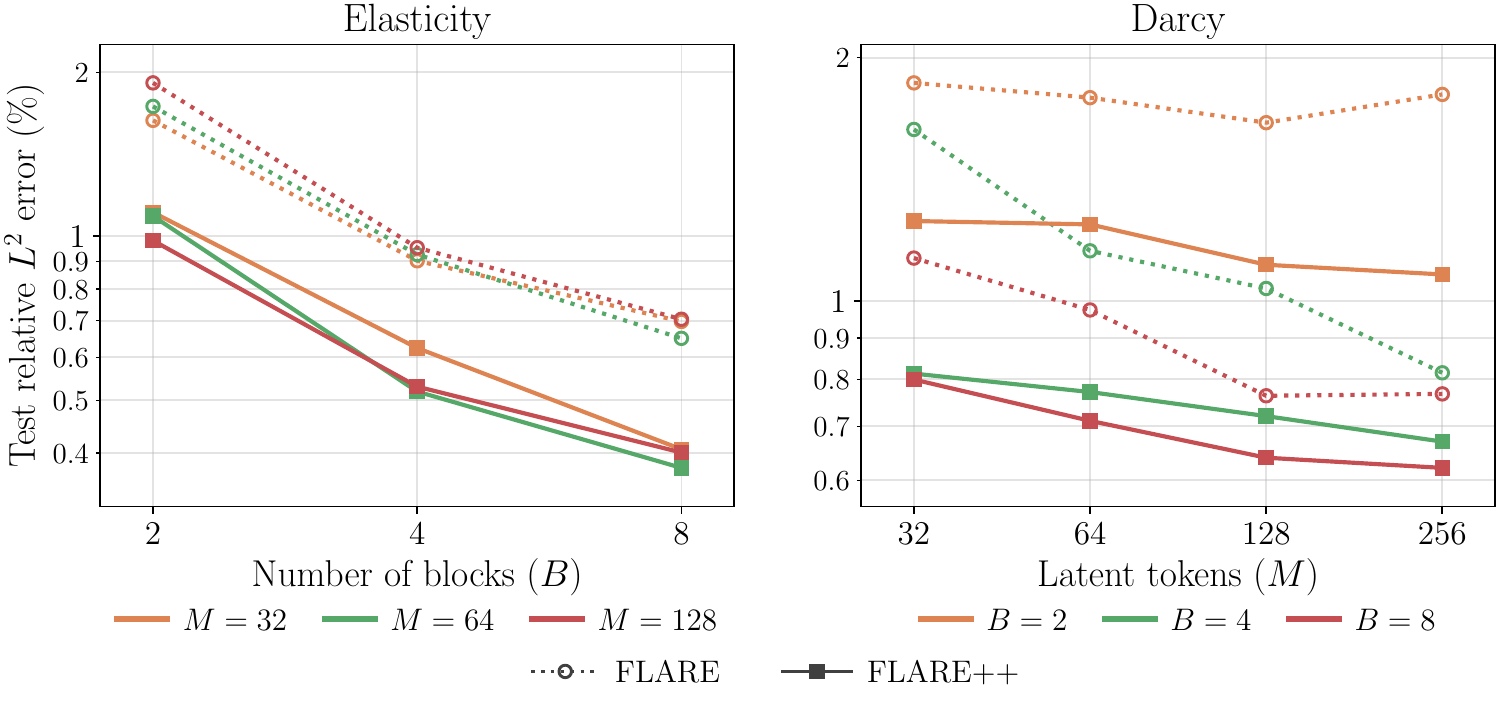}
\caption{
Test error of FLARE and FLARE++ on Elasticity and Darcy over the joint latent-budget and depth grid $(M,B)$.
FLARE++ is more accurate in every cell; on Darcy both mixers benefit from additional latents, with diminishing returns beyond $M=128$ at $B=8$.
}
\label{fig:abl_grid}
\end{figure}

\subsection{Standard PDE surrogate benchmarks}
\label{sec:experiments_pde_surrogates}

\paragraph{Benchmark problems.}
We evaluate on the Elasticity, Darcy, Airfoil, and Pipe benchmarks of the Transolver and LNO papers~\citep{wu2024transolver,wang2024latent} and on the DrivAerML-40K benchmark of the FLARE study~\citep{puri2026flare}, spanning structured and unstructured discretizations of $1$K--$40$K points (\autoref{fig:gallery_2d}, \autoref{app:benchmark_datasets}).

\paragraph{Comparison protocol.}
We compare FLARE++ with FLARE, the Transolver family, Luna, AB-UPT, and full self-attention as token mixers in a shared residual backbone~\citep{puri2026flare,wu2024transolver,luotransolver++,zhou2026transolver3,ma2021luna,alkin2025abuptscalingneuralcfd,vaswani2017attention}: channel width, head count, depth, latent budget, input and output projections, feed-forward blocks, and training settings are fixed within each dataset, so the comparison isolates the token mixer (\autoref{app:comparison_model_construction}).
AB-UPT uses $512$ anchors throughout, since at the $64$--$128$ latent budgets of the other mixers its error is markedly worse.
Parameter counts therefore differ by mixer (\autoref{tab:param_counts}), which is also why our FLARE entries are not those of the parameter-matched table in the FLARE study (\autoref{app:comparison_model_construction}); FLARE++ carries $14$--$23\%$ more parameters than FLARE, and we account for the mixer's cost in step time and memory at a given token count (\autoref{sec:experiments_efficiency}) and through the joint depth and latent-budget sweeps (\autoref{sec:experiments_ablations}).
Set Transformer is evaluated as a complete architecture; PerceiverIO, GNOT, and LNO results are reproduced from the FLARE study~\citep{lee2019settransformer,jaegle2021perceiverio,hao2023gnot,wang2024latent}.
All runs use FP32 and report test relative $L^2$ error; full self-attention provides a quadratic-cost reference on the three smallest problems.
Since \autoref{sec:drivaerml_scaling} uses FP16, \autoref{sec:experiments_precision} checks that precision does not alter the mixer ranking.

\paragraph{Results.}
FLARE++ achieves the lowest error among the efficient models compared on all five benchmarks (\autoref{tab:pde_benchmark_summary}).
It reduces FLARE's error by $12$--$45\%$ ($25\%$ on average) and Transolver-3's by $20$--$43\%$ ($31\%$ on average).
Averages are over per-task relative reductions.
The improvement spans solid mechanics, porous flow, and aerodynamics.
The gains extend beyond comparisons with PDE-specific architectures.
FLARE++ improves on Set Transformer and on Luna on all five tasks.
The results favor adapting a compact set of routes to the field.
Full self-attention provides a complementary reference for the cost of compression.
FLARE++ has lower error on Elasticity and Airfoil, while full self-attention retains the lowest Darcy error.
A low-rank routing bottleneck can therefore support accurate predictions without reproducing unrestricted token-to-token attention.

\subsection{Industrial-scale aerodynamics: DrivAerML}
\label{sec:drivaerml_scaling}

\begin{wrapfigure}{r}{0.7\linewidth}
\centering
\vspace{-1.4em}
\includegraphics[width=\linewidth]{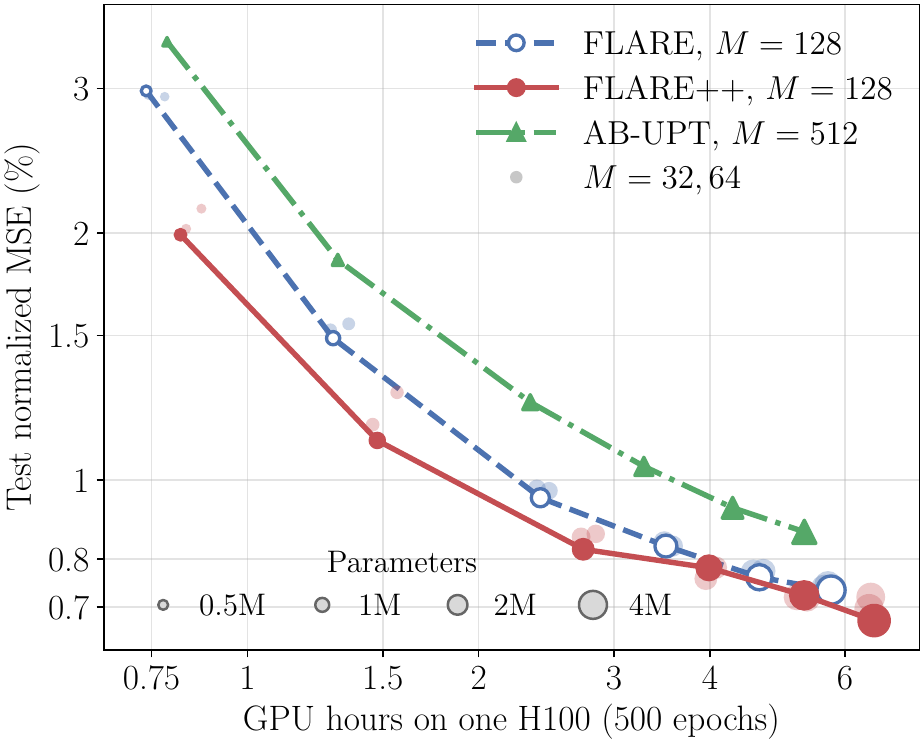}
\caption{
Test normalized MSE on full-surface DrivAerML against GPU hours per training run on one NVIDIA H100.
Lines trace depth at $M=128$ for FLARE and FLARE++ and $512$ anchors for AB-UPT.
Marker area is proportional to the parameter count, $0.37$--$4.72$M over $B\in\{2,4,8,12,16,20\}$.
\autoref{fig:drivaerml_scaling} plots the runs against depth.
}
\label{fig:drivaerml_gpu_hours}
\vspace{-1.2em}
\end{wrapfigure}

\paragraph{Setup.}
DrivAerML~\citep{ashton2024drivaerml} tests whether the routing gains persist in industrial-scale automotive aerodynamics (\autoref{fig:gallery_drivaerml}).
Models predict surface pressure and three wall-shear components, training on $100$K sampled cells per step and evaluating on the full $8.2$M-cell surface under context parallelism.
We compare FLARE and FLARE++ across three latent budgets and six depths with the same backbone and training settings, and include the surface-only AB-UPT mixer with $512$ anchors.
We also attempted to include Transolver-3, but its official recipe is FP32 only and every 16-bit run we tried on DrivAerML diverged (\autoref{app:comparison_model_construction}).
The metric is normalized four-channel MSE (in \%), distinct from the pressure-only relative $L^2$ error of DrivAerML-40K.
We report it against training cost, measured as the GPU hours to train each run for $500$ epochs on one NVIDIA H100 (\autoref{app:drivaerml_sweep}).
\autoref{app:drivaerml_discussion} gives the sampling, normalization, and model details.

\paragraph{Results.}
FLARE++ improves on FLARE in all $18$ matched configurations of depth and latent budget, by $31\%$ at two blocks, $22\%$ at four, and $13\%$ at eight when averaged over latent budgets, and by $6$--$7\%$ in the deepest models tested (\autoref{app:drivaerml_sweep}).
A full FLARE++ training run costs $1.1$--$1.2\times$ the GPU hours of the matching FLARE run (\autoref{fig:drivaerml_gpu_hours}).
Above $3.4$ GPU hours, every FLARE run is dominated by a FLARE++ run that is both cheaper and more accurate, and FLARE's lowest error, $0.734\%$ at $5.75$ GPU hours, is undercut by FLARE++ at $5.19$ GPU hours.
Below $2.5$ GPU hours the two models alternate, although at the smallest budget FLARE++ lowers the error from $2.93$--$2.98\%$ to $1.99\%$ for under $0.1$ additional GPU hours.
AB-UPT has higher error than FLARE and FLARE++ at every depth despite its larger anchor budget, and every AB-UPT run is dominated by a FLARE or FLARE++ run of equal or lower cost.
Together with the standard benchmarks, these results show that query synthesis improves prediction across a wide range of spatial resolutions.

\subsection{Long Range Arena benchmark}
\label{sec:experiments_lra}

We evaluate FLARE++ on Long Range Arena (LRA)~\citep{taylong} with FLARE's backbone and task configurations~\citep{puri2026flare}, to test whether the routing improvement transfers to sequence and image tasks.
FLARE++ improves on FLARE on all five tasks, raising average accuracy from $58.08\%$ to $61.60\%$, second only to the published Luna result ($61.95\%$), and leads on two tasks.
\autoref{app:lra} reports full results and protocol.

\section{Scaling, efficiency, and ablations}
\label{sec:analysis}

\subsection{Latent budget and residual depth}
\label{sec:experiments_ablations}

We jointly vary the latent budget $M$ and residual depth $B$ of FLARE and FLARE++ on Elasticity and Darcy, with other settings as in \autoref{app:standard_pde_configs} (\autoref{fig:abl_grid}).
FLARE++ is more accurate than FLARE in all $21$ matched configurations, by $16$--$50\%$, so the gain persists across communication budgets and backbone depths.
At a fixed latent budget, FLARE++ reaches a given error at a shallower depth than FLARE on both tasks.
Increasing the latent budget alone does not recover this accuracy: on Elasticity, raising $M$ from $32$ to $128$ leaves FLARE unchanged at every depth while FLARE++ improves (marginally at $B=8$), so the same rank budget predicts better with adaptive routing queries.

\subsection{Efficiency and scaling}
\label{sec:experiments_efficiency}

\begin{figure}[!t]
\centering
\includegraphics[width=\linewidth]{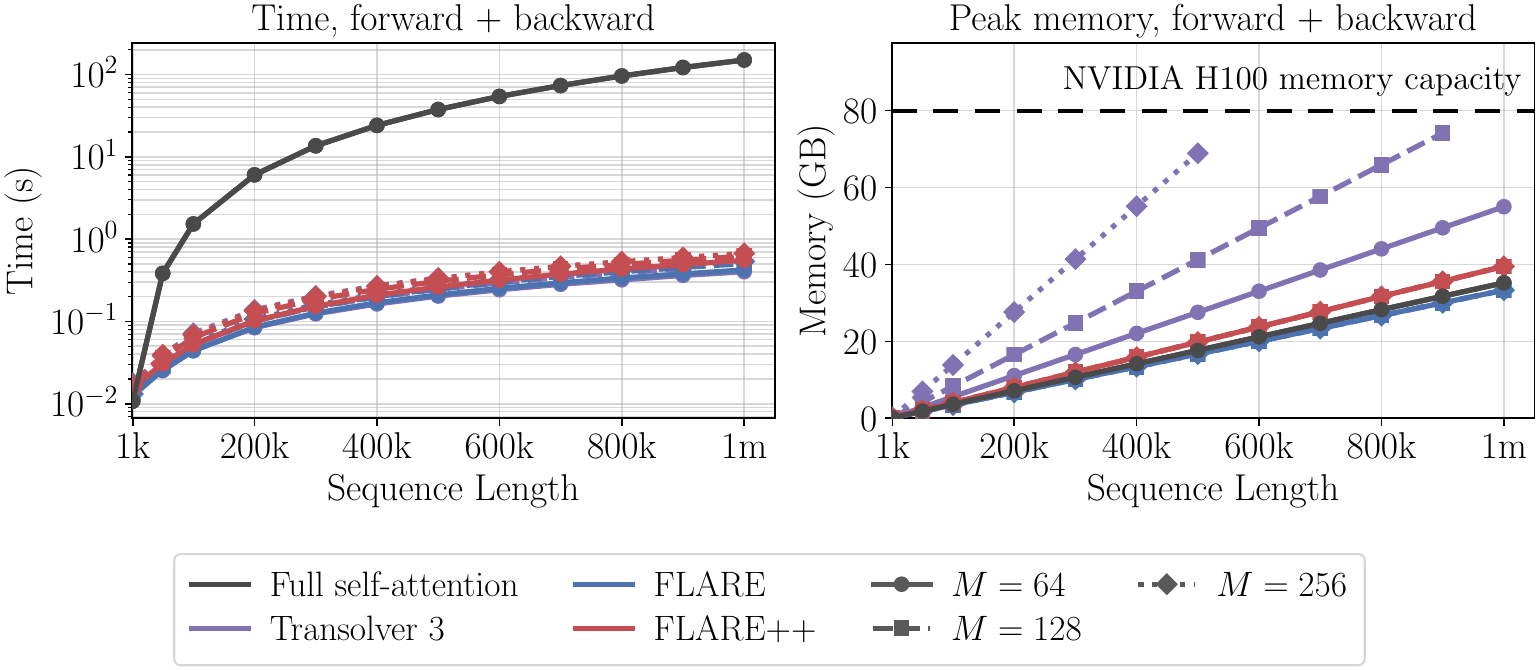}
\caption{
Forward-plus-backward time and peak memory against input size at $B=8$, $C=128$, $H=8$, FP16, on one NVIDIA H100; \autoref{app:time_memory} gives the measurement details.
}
\label{fig:time_memory}
\end{figure}

On a single H100, from $1$K to $1$M tokens (\autoref{fig:time_memory}, \autoref{app:time_memory}), fused SDPA keeps memory linear in $N$ for every attention model tested, but full self-attention retains quadratic work and at $500$K tokens is two orders of magnitude slower than FLARE++.
FLARE++ costs $1.2$--$1.3\times$ FLARE's step time and $1.18\times$ its peak memory, stable beyond $100$K tokens and below the $4/3$--$3/2$ mixer-arithmetic increase of \autoref{sec:flarepp} because the feed-forward network dilutes it.
On up to four H100 GPUs, context parallelism keeps time and memory efficiencies at or above about $0.91$ and $0.95$ (\autoref{app:multigpu_scaling}).

\subsection{Mixed-precision sensitivity}
\label{sec:experiments_precision}

\autoref{app:precision} repeats the depth sweep in FP32 and FP16 for full self-attention, FLARE, and FLARE++ on Elasticity, Darcy, and Airfoil.
Precision changes full self-attention's error by up to $82\%$, but FLARE++ remains below FLARE in every comparison and is the least sensitive mixer: its FP16 error stays within $8\%$ of its FP32 error at every depth, against $0.88$--$1.33\times$ for FLARE and $0.70$--$1.82\times$ for full self-attention.
Precision thus confounds comparisons against full self-attention but not between latent mixers; FLARE++'s stability comes from its fixed reference and gate (\autoref{app:flarepp_blended}).

\section{Conclusion}

FLARE++ lets a low-rank attention operator decide from the input itself how to compress and redistribute it, rather than reusing queries fixed at training time, with FLARE's attention primitive, at linear cost, on meshes too large for one GPU.
It improves on fixed-query FLARE across PDE benchmarks, full-surface DrivAerML, and Long Range Arena at $1.1$--$1.2\times$ FLARE's time per full training run.
The depth and latent-budget sweeps indicate that improved routing, not added depth or latents, is what reaches a given accuracy with a shallower model.

\subsection*{AI use statement}

We used generative AI tools (large language model assistants) throughout this work.
They were used to brainstorm and design the method, to formulate and check the mathematical claims, to write and run the experiment, analysis, and plotting code, to analyze the results, and to draft and edit the text of this paper and format its references.
We did not use generative AI tools to generate data; all datasets are the published benchmarks cited in the paper.
All AI-assisted work was reviewed by the authors: the derivations were checked by hand, the context-parallel implementation was tested numerically against its single-device counterpart, every reported number was traced to its experiment logs, and all text was revised by the authors.
We take responsibility for the final content of this work, including the method, claims, code, and text produced with the aid of generative AI.

\subsection*{Reproducibility statement}

\autoref{app:mixer_ops} specifies the mixer operations, and \autoref{app:multigpu} specifies the context-parallel implementation and its backward pass.
The remaining appendices document experiment configurations, extended results, and qualitative predictions.
The training code and runnable scripts for the key results are available at \url{https://github.com/vpuri3/FLARE.py}.

\bibliographystyle{iclr2027_conference}
\bibliography{main}

\appendix

\section{FLARE++ mixer architecture}
\label{app:mixer_ops}

This appendix records the design of the FLARE++ mixer in more detail than \autoref{sec:flarepp}: why synthesized routing queries are hard to optimize, how normalization and the fixed reference stabilize them, the near-identity initialization of the feature projections, and the shared synthesis keys and values.
\autoref{tab:mixer_ops} lists every operation of FLARE, FLARE++, and the simplified variant in execution order.
The simplified variant drops the fixed reference and gate and routes with $Q^{\mathrm d}_h(X)$ alone; \autoref{app:simplified_flarepp} defines it and reports the ablation.
Throughout, heads are treated independently and $D=C/H$ is the head width.

\subsection{Query synthesis}
FLARE routes every token through $M\ll N$ latent queries that are learned once and held fixed across samples.
FLARE++ instead adapts the routing queries to the current input: with learned seeds $\widetilde Q_h\in\mathbb{R}^{M\times D}$, learned fixed queries $Q^{\mathrm f}_h\in\mathbb{R}^{M\times D}$, a per-head gate $g_h$, and projections $\widetilde K_h,\widetilde V_h,K_h,V_h$ of $X$, one block computes
\begin{align}
    Q^{\mathrm d}_h&=\operatorname{SDPA}(\widetilde Q_h,\widetilde K_h,\widetilde V_h),&
    Q_h&=\mathcal N_0(Q^{\mathrm f}_h)+g_h\,Q^{\mathrm d}_h,\notag\\
\label{eq:app_flarepp_three_calls}
    Z_h&=\operatorname{SDPA}(Q_h,K_h,V_h),&
    Y_h&=\operatorname{SDPA}(K_h,Q_h,Z_h),
\end{align}
as in \eqref{eq:flarepp_dynamic_query}--\eqref{eq:flarepp_mixer}, with $\widetilde V_h=\widetilde K_h$ in the default configuration (\autoref{app:flarepp_shared_kv}) and with the per-head normalizations of \autoref{sec:flarepp} introduced below one at a time.
The first call synthesizes an input-dependent correction to the $M$ routing vectors; the second gathers the input into $M$ latents through them, and the third scatters the latents back to the $N$ input locations.
This appendix keeps the general form with a separate $\widetilde V_h$, since the simplified variant of \autoref{app:simplified_flarepp} uses it.

\subsection{Optimization difficulty of synthesized routing queries}
\label{app:flarepp_difficulty}
The principal optimization challenge in FLARE++ is that the output of the first attention call is the query of the second.
Writing the synthesis attention as
\begin{equation}
\label{eq:app_synthesis_attention}
    Q^{\mathrm d}_h=\widetilde A_h\widetilde V_h,
    \qquad
    \widetilde A_h=\operatorname{softmax}\!\left(\frac{\widetilde Q_h\widetilde K_h^\top}{\sqrt D}\right),
\end{equation}
the gather weights of the encoder are $A_h=\operatorname{softmax}(Q_hK_h^\top/\sqrt D)$, so updates to the synthesis branch also change the scores used by the value-gathering encoder.

This is a difficult problem at initialization.
The seeds $\widetilde Q_h$ and the projections producing $\widetilde K_h$ and $\widetilde V_h$ initially carry little task-specific structure, so $\widetilde A_h$ is weakly structured and often close to diffuse, while the values it aggregates have not yet learned a meaningful basis.
The synthesized queries
\begin{equation}
\label{eq:app_synthesized_query_rows}
    Q^{\mathrm d}_{h,m}=\sum_{n=1}^{N}\widetilde A_{h,mn}\,\widetilde V_{h,n}
\end{equation}
can therefore change substantially from one optimization step to the next early in training.
This nonstationarity matters because $Q^{\mathrm d}_h$ is not merely an intermediate feature: it enters the routing query $Q_h$, whose direction sets the next attention logits through $Q_hK_h^\top$, so a change in the first call changes the coordinate system in which the second call routes.
The encoder must learn useful routing directions while the routing vectors themselves are moving.
We found that controlling this early-training instability explicitly is important for optimizing FLARE++ reliably.
FLARE++ does so in two ways: it normalizes the inputs of the synthesis call, and it blends the synthesized correction with a fixed learned reference.

\subsection{Normalizing the synthesis call}
\label{app:flarepp_normalized}
The first stabilization applies query--key normalization~\citep{henry2020qknorm,dehghani2023vit22b} to the inputs of the synthesis call,
\begin{equation}
\label{eq:app_qk_norm}
    Q^{\mathrm d}_h=\operatorname{SDPA}\!\big(\mathcal N_a(\widetilde Q_h),\mathcal N_0(\widetilde K_h),\widetilde V_h\big),
\end{equation}
so that the synthesis logits $\widetilde S_{h,mn}=\mathcal N_a(\widetilde Q_h)_m^\top\mathcal N_0(\widetilde K_h)_n/\sqrt D$ no longer inherit uncontrolled variation in the magnitudes of the seeds and keys.
The seeds carry a learned scale, which sets the temperature of the synthesis logits; the keys do not, for the reason given in \autoref{app:flarepp_blended}.
Early changes in the attention distribution are then driven by changes in feature direction rather than by simultaneous changes in direction and scale.
We use this normalization in all reported FLARE++ and simplified FLARE++ runs to stabilize the synthesis logits.
The main text writes RMSNorm~\citep{zhang2019root} throughout; in full precision the same placement uses LayerNorm, whose affine-free output also has unit RMS, so the bounds below are unchanged.
LayerNorm was markedly more sensitive to FP16 arithmetic in this setting, whereas RMSNorm was reliable, so every mixed-precision run uses RMSNorm.
The normalization acts on the inputs to the synthesis call and not on its output $Q^{\mathrm d}_h$: as \autoref{app:flarepp_blended} argues, the magnitude of a synthesized query carries information that should not be discarded.

\subsection{Fixed reference and gate}
\label{app:flarepp_blended}
The second stabilization keeps FLARE's learned routing basis and lets the synthesis call learn an input-dependent correction to it.
FLARE++ combines the synthesized query $Q^{\mathrm d}_h(X)$ with an independent learned set of fixed queries $Q^{\mathrm f}_h\in\mathbb{R}^{M\times D}$,
\begin{equation}
\label{eq:app_blend}
    Q_h(X)=\mathcal N_0(Q^{\mathrm f}_h)+g_h\,Q^{\mathrm d}_h(X),
    \qquad
    g_h=\sigma(\gamma_h)\in(0,1),
\end{equation}
as in \eqref{eq:flarepp_anchored_query}.
The gate holds one learned scalar $\gamma_h$ per head, broadcast over latent queries and channels.
In the limit $g_h\to0$, the queries become input-independent; the normalization choices still differ from FLARE.
We also trained FLARE with the same query--key normalization applied to its fixed queries and routing keys and saw no improvement.
Increasing $g_h$ admits a larger input-dependent correction.
We initialize $\gamma_h=0.25$, so $g_h=\sigma(0.25)\approx0.56$: the model starts with a substantial synthesized contribution rather than collapsing to FLARE, while the fixed component provides a stable routing anchor from the first step.
The blend is evaluated in FP32 and cast back to the model datatype, which costs $\mathcal{O}(MC)$ per block against the $\mathcal{O}(NC^2)$ projections and $\mathcal{O}(NMC)$ attention calls, and so adds negligible overhead.

\paragraph{Why the synthesized query is not normalized.}
A natural alternative is to normalize $Q^{\mathrm d}_h$ separately before blending it with $Q^{\mathrm f}_h$.
We found this undesirable.
By \eqref{eq:app_synthesized_query_rows}, each synthesized query is a convex combination of the value rows selected by the synthesis attention.
Early in training both $\widetilde A_h$ and $\widetilde V_h$ evolve rapidly, so both the direction and the magnitude of $Q^{\mathrm d}_h$ vary considerably, and normalizing the aggregate would remove the magnitude entirely.
A small magnitude need not signal a numerical failure: it arises naturally when the attended value vectors point in inconsistent directions and partially cancel, while a large magnitude arises when they agree.
The magnitude of $Q^{\mathrm d}_h$ therefore encodes how coherently the input supports a routing direction, and normalizing it would rescale a weak or poorly aligned correction to the same nominal size as a strongly supported one, converting a weak routing signal into a full-strength update.
We instead control the scale of $Q^{\mathrm d}_h$ before aggregation, by normalizing the individual value vectors.

\paragraph{Bounding the synthesized query through value normalization.}
FLARE++ applies an affine-free RMS normalization $\mathcal N_0$, with no learned scale or bias, to the synthesis values, so that
\begin{equation}
\label{eq:app_value_norm}
    Q^{\mathrm d}_{h,m}=\sum_{n=1}^{N}\widetilde A_{h,mn}\,\mathcal N_0(\widetilde V_h)_n,
    \qquad
    \operatorname{RMS}\big(\mathcal N_0(\widetilde V_h)_n\big)=1.
\end{equation}
Since every row of $\widetilde A_h$ is a probability distribution and $\operatorname{RMS}(\cdot)=\|\cdot\|_2/\sqrt D$ is a norm up to a constant, the triangle inequality gives
\begin{equation}
\label{eq:app_rms_bound}
    \operatorname{RMS}\big(Q^{\mathrm d}_{h,m}\big)
    \le\sum_{n=1}^{N}\widetilde A_{h,mn}\operatorname{RMS}\big(\mathcal N_0(\widetilde V_h)_n\big)=1,
    \qquad\text{equivalently}\qquad
    \big\|Q^{\mathrm d}_{h,m}\big\|_2\le\sqrt D.
\end{equation}
This gives the desired scale control without normalizing the aggregated query.
The distinction matters: normalizing $Q^{\mathrm d}_h$ would force every latent correction to the same magnitude, whereas normalizing the constituent values only imposes an upper bound on the convex aggregate, and $Q^{\mathrm d}_h$ remains free to be small when the attended directions disagree.
The fixed component is normalized by the same affine-free $\mathcal N_0$ for the complementary reason: its role is a stable anchor, and without a learned scale its RMS is exactly one throughout training.
Combining the two, every row of the routing query in \eqref{eq:app_blend} satisfies $1-g_h\le\operatorname{RMS}(Q_{h,m})\le1+g_h$, with $g_h<1$.
The blend thus pairs a controlled fixed reference with a bounded but magnitude-preserving displacement, and can be read as an input-dependent deformation of FLARE's latent routing basis in which each head adapts its routing directions to the current input while staying anchored to a stable learned coordinate system.
The routing keys $K_h=\mathcal N_a(XW_{K,h})$ are normalized with a learned scale: with the query magnitude controlled as above, that scale sets the temperature of the routing logits $Q_hK_h^\top/\sqrt D$, as the seed scale does for the synthesis logits.

\subsection{Near-identity feature projections}
\label{app:flarepp_identity_init}
A second optimization choice concerns the projections producing $\widetilde K_h$, $\widetilde V_h$, $K_h$, and $V_h$.
Rather than arbitrary random linear maps, we initialize them close to the identity,
\begin{equation}
\label{eq:app_identity_init}
    W=I+\Delta W,
    \qquad
    \Delta W_{ij}\sim\operatorname{TruncNormal}(0,\,0.02),
\end{equation}
with zero biases, so that $K_h\approx X_h$ and $\widetilde K_h\approx X_h$ at initialization, and likewise for the value projections.
All three mixers use this initialization for every feature projection.
This initialization is particularly useful for FLARE++ because the quality of the synthesized queries depends on the representation produced by an upstream attention call.
With a conventional random projection, the encoder initially sees routing keys $K_h$ in an arbitrary rotated and rescaled basis while the synthesis call tries to learn useful queries in another, independently evolving basis.
Near-identity initialization preserves the input representation at the start of training, so the routing score $Q_hK_h^\top\approx Q_hX_h^\top$ provides a direct learning signal between the synthesized queries and the existing input features from the first optimization steps.
The same holds for $\widetilde K_h\approx X_h$, which lets the synthesis call begin by selecting among meaningful input features rather than among the outputs of an unrelated random transformation.
The projections remain fully learnable, so the initialization restricts nothing about the final representation; it only provides a well-conditioned starting point.

\subsection{Sharing synthesis keys and values}
\label{app:flarepp_shared_kv}
We find little or no loss in accuracy from sharing the synthesis keys and values, $\widetilde V_h=\widetilde K_h$, under which
\begin{equation}
\label{eq:app_shared_kv}
    Q^{\mathrm d}_h=\operatorname{softmax}\!\left(\frac{\mathcal N_a(\widetilde Q_h)\,\mathcal N_0(\widetilde K_h)^\top}{\sqrt D}\right)\mathcal N_0(\widetilde K_h).
\end{equation}
Each synthesized query is then an attention-weighted convex combination of normalized input feature directions, and its magnitude reflects their directional agreement, which further motivates preserving the magnitude of $Q^{\mathrm d}_h$ rather than renormalizing it after aggregation.
The additional projection, seeds, and gates increase the mixer parameter count from FLARE's $3C^2+MC$ to $4C^2+2MC+H$ per block, excluding biases and normalization parameters.
Sharing removes one $C\to C$ projection, saving $\mathcal{O}(NC^2)$ computation and $\mathcal{O}(NC)$ activation storage per block while preserving the behaviour of FLARE++ in our experiments, so shared synthesis keys and values are the default FLARE++ configuration.

\subsection{Operations in one block}
\autoref{tab:mixer_ops} lists, in execution order, every projection, normalization, and attention call inside one block of FLARE, FLARE++, and simplified FLARE++ at the default mixed-precision settings used for every FP16 run in the paper, taken from the reference implementation.
All three share the block structure around the mixer: a pre-normalization of the residual stream, the mixer, a residual addition, and a normalized feed-forward update.
The differences are confined to the routing scores: how the routing queries are formed and which per-head query and key tensors are normalized; the value, output, and feed-forward rows are identical.
Simplified FLARE++ normalizes only the synthesis seeds and keys; FLARE++ also normalizes the fixed query and the routing keys, shares the synthesis values with the synthesis keys, and accumulates its gated blend in FP32.
Each per-head normalization acts on the head dimension $D$ and carries $D$ parameters when it is affine.

\begin{table}[t]
\centering
\caption{
Operations in one mixer block, in execution order, for the three mixers in mixed precision with RMSNorm.
$X$ is the block input, $\mathcal N_a$ an RMSNorm over the head dimension with a learned scale, $\mathcal N_0$ one without, and $\sigma$ the logistic function; ``learned'' marks a free parameter tensor and ``--'' an operation the mixer does not perform.
Every projection is a plain linear map initialized near the identity (\autoref{app:flarepp_identity_init}).
The gate logits $\gamma_h$ are initialized at $0.25$, so every gate starts at $\sigma(0.25)\approx0.56$.
Parameter counts exclude biases and normalization scales.
}
\label{tab:mixer_ops}
\small
\setlength{\tabcolsep}{5pt}
\renewcommand{\arraystretch}{1.15}
\begin{tabular}{l l l l}
\toprule
\textbf{Operation} & \textbf{FLARE} & \textbf{FLARE++} & \textbf{Simplified FLARE++} \\
\midrule
Block input & $X\leftarrow\mathrm{RMSNorm}(X)$ & same & same \\
\midrule
Synthesis seeds $\widetilde Q_h$ & -- & learned, $\mathcal N_a$ & learned, $\mathcal N_a$ \\
Synthesis keys $\widetilde K_h$ & -- & $X\widetilde W_{K,h}$, $\mathcal N_0$ & $X\widetilde W_{K,h}$, $\mathcal N_a$ \\
Synthesis values $\widetilde V_h$ & -- & shared, $\widetilde V_h=\widetilde K_h$ & $X\widetilde W_{V,h}$, no norm \\
Synthesized query $Q^{\mathrm d}_h$ & -- & $\operatorname{SDPA}(\widetilde Q_h,\widetilde K_h,\widetilde V_h)$ & $\operatorname{SDPA}(\widetilde Q_h,\widetilde K_h,\widetilde V_h)$ \\
Fixed query $Q^{\mathrm f}_h$ & learned, no norm & learned, $\mathcal N_0$ & -- \\
Routing query $Q_h$ & $Q^{\mathrm f}_h$ & $Q^{\mathrm f}_h+\sigma(\gamma_h)\,Q^{\mathrm d}_h$ in FP32 & $Q^{\mathrm d}_h$, no norm \\
\midrule
Routing keys $K_h$ & $XW_{K,h}$, no norm & $XW_{K,h}$, $\mathcal N_a$ & $XW_{K,h}$, no norm \\
Values $V_h$ & $XW_{V,h}$, no norm & same & same \\
Encoder & $Z_h=\operatorname{SDPA}(Q_h,K_h,V_h)$ & same & same \\
Decoder & $Y_h=\operatorname{SDPA}(K_h,Q_h,Z_h)$ & same & same \\
Output & $YW_O$ & same & same \\
\midrule
Feed-forward & $X+\mathrm{FFN}(\mathrm{RMSNorm}(X))$ & same & same \\
\midrule
Projections & $3$ & $4$ & $5$ \\
SDPA calls & $2$ & $3$ & $3$ \\
Per-head norms & $0$ & $4$ & $2$ \\
Learned latents & $Q^{\mathrm f}$ & $Q^{\mathrm f}$, $\widetilde Q$, $\gamma$ & $\widetilde Q$ \\
Mixer parameters & $3C^2+MC$ & $4C^2+2MC+H$ & $5C^2+MC$ \\
\bottomrule
\end{tabular}
\end{table}

\subsection{Simplified FLARE++}
\label{app:simplified_flarepp}

The simplified variant routes with the synthesized query alone,
\begin{equation}
\label{eq:app_simplified}
    Q_h(X)=Q^{\mathrm d}_h(X)=\operatorname{SDPA}\!\big(\mathcal N_a(\widetilde Q_h),\mathcal N_a(\widetilde K_h),\widetilde V_h\big),
\end{equation}
with no fixed reference, no gate, a separate and unnormalized value projection $\widetilde V_h=X\widetilde W_{V,h}$, a learned scale on the synthesis keys, and no normalization of the routing keys (\autoref{tab:mixer_ops}).
Its routing query therefore carries no magnitude bound.
It has $5C^2+MC$ mixer parameters and costs $\mathcal{O}(N(5C^2+3MC))$ per block.
It was our first formulation, and we keep it as the ablation that isolates what the fixed reference adds.

\paragraph{Accuracy.}
\autoref{tab:simplified_benchmark} repeats the standard-benchmark comparison of \autoref{tab:pde_benchmark_summary} for both variants.
In FP32 the two are interchangeable.
The simplified variant is lower on Darcy and Pipe and FLARE++ on Elasticity, Airfoil, and DrivAerML-40K, and averaged over the five benchmarks the simplified variant reduces FLARE's error by $24\%$ and FLARE++ by $25\%$.
On the $(M,B)$ grid of \autoref{sec:experiments_ablations}, both improve on FLARE in all $21$ cells; the simplified variant is lower in eight of the nine Elasticity cells, by $3$--$13\%$, and the two split the twelve Darcy cells evenly.

\begin{table}[ht]
\centering
\caption{
Test relative $L^2$ error (in \%) of FLARE, FLARE++, and simplified FLARE++ on the standard PDE benchmarks under the protocol of \autoref{tab:pde_benchmark_summary}; single seed, FP32.
}
\label{tab:simplified_benchmark}
\setlength{\tabcolsep}{6pt}
\begin{tabular}{lccccc}
\toprule
\textbf{Model} & \textbf{Elasticity} & \textbf{Darcy} & \textbf{Airfoil} & \textbf{Pipe} & \textbf{DrivAerML-40K} \\
\midrule
FLARE & 0.64 & 0.76 & 0.57 & 0.51 & 7.22 \\
Simplified FLARE++ & 0.38 & 0.59 & 0.52 & 0.34 & 6.04 \\
FLARE++ & 0.37 & 0.62 & 0.50 & 0.35 & 5.80 \\
\bottomrule
\end{tabular}
\end{table}

\paragraph{Why the fixed reference.}
Three properties separate the variants where the small FP32 benchmarks do not.
First, FLARE++ contains FLARE as the $g_h\to0$ limit and starts training from a stable learned routing basis, whereas the simplified variant must learn useful routing directions while its only routing vectors are moving (\autoref{app:flarepp_difficulty}).
Second, FLARE++ is far less sensitive to mixed precision.
In the precision sweep of \autoref{app:precision}, its FP16 error stays within $8\%$ of its FP32 error on Elasticity, Darcy, and Airfoil, against up to $23\%$ for the simplified variant; on Pipe, the simplified variant's FP16 error reaches $3.5\times$, $4.8\times$, and $12.9\times$ its FP32 error at $B=2,4,8$ while FLARE stays within $1.4\times$.
Every large-scale run in the paper is in FP16, so this property decides which variant can be used at scale.
Third, sharing the synthesis keys and values removes one projection: under the protocol of \autoref{fig:time_memory}, FLARE++ costs $1.2$--$1.3\times$ FLARE's step time against $1.25$--$1.5\times$ for the simplified variant, $12$--$20\%$ less at $M=64$, at the same peak memory.

\section{Multi-GPU context parallelism}
\label{app:multigpu}

\begin{table}[H]
\centering
\caption{
Cost of the distributed FLARE encoder, per rank and per invocation, in the accounting of \autoref{fig:flarepp_block} ($C=HD$, so $HMD=MC$).
Communication counts the values a rank contributes to a collective; local compute counts arithmetic on tensors the rank already holds.
The encoder collectives communicate only latent quantities: $\bar K_r$ and $\bar V_r$ stay on the shard that owns those keys and values.
The accounting excludes synchronization of replicated projection and backbone parameter gradients.
}
\label{tab:merge_cost}
\footnotesize
\setlength{\tabcolsep}{5pt}

\begin{subtable}{\linewidth}
\centering
\caption{
Forward pass.
The first row, the fused local attention over the rank's own shard, is the only $N$-dependent entry.
}
\label{tab:merge_forward_cost}
\begin{tabular}{@{}llll@{}}
\toprule
\textbf{Step} & \textbf{Collective} & \textbf{Communication} & \textbf{Local compute} \\
\midrule
$Z_r,L_r=\operatorname{SDPA}(Q,K_r,V_r)$ & --- & --- & $\mathcal{O}(N_rMHD)$ \\
\midrule
$L_{\max}=\max_r L_r$        & all-reduce max & $\mathcal{O}(HM)$  & --- \\
$d_r=\exp(L_r-L_{\max})$     & ---            & ---                & $\mathcal{O}(HM)$ \\
$U_r=d_r Z_r$                & ---            & ---                & $\mathcal{O}(HMD)$ \\
$d=\sum_r d_r$               & all-reduce sum & $\mathcal{O}(HM)$  & --- \\
$U=\sum_r U_r$               & all-reduce sum & $\mathcal{O}(HMD)$ & --- \\
$Z=U/d$                      & ---            & ---                & $\mathcal{O}(HMD)$ \\
$L=L_{\max}+\log d$          & ---            & ---                & $\mathcal{O}(HM)$ \\
\midrule
Encoder total                & 3 collectives  & $\mathcal{O}(HM(D\!+\!2))$ & $\mathcal{O}(N_rMHD)$ \\
\bottomrule
\end{tabular}
\end{subtable}

\vspace{1.25em}

\begin{subtable}{\linewidth}
\centering
\caption{
Backward pass for the general two-output derivative; the zero log-normalizer cotangent needs no reduction in FLARE or FLARE++.
The upper block differentiates the merge; the lower block denotes a local attention backward accepting cotangents of both the output and log-normalizer.
It returns $\bar K_r$, $\bar V_r$, and $\bar Q^{(r)}$ without materializing $S_r$; the text describes fused-backward reuse when $\bar L=0$.
}
\label{tab:merge_backward_cost}
\begin{tabular}{@{}llll@{}}
\toprule
\textbf{Step} & \textbf{Collective} & \textbf{Communication} & \textbf{Local compute} \\
\midrule
$\bar Z=\sum_r\bar Z^{(r)}$ & all-reduce sum & $\mathcal{O}(HMD)$ & --- \\
$\bar L=\sum_r\bar L^{(r)}$ & all-reduce sum & $\mathcal{O}(HM)$  & --- \\
$\bar Z_r=\exp(L_r-L)\,\bar Z$ & --- & --- & $\mathcal{O}(HMD)$ \\
$\bar L_r=\exp(L_r-L)\left(\bar L+\left\langle\bar Z,Z_r-Z\right\rangle_D\right)$ & --- & --- & $\mathcal{O}(HMD)$ \\
\midrule
$\bar Q^{(r)},\bar K_r,\bar V_r=\operatorname{SDPA}_{\mathrm{bwd}}(\bar Z_r,\bar L_r)$ & --- & --- & $\mathcal{O}(N_rMHD)$ \\
$\bar Q=\sum_r\bar Q^{(r)}$ & all-reduce sum & $\mathcal{O}(HMD)$ & --- \\
\midrule
Encoder total & 3 collectives & $\mathcal{O}(HM(2D\!+\!1))$ & $\mathcal{O}(N_rMHD)$ \\
\bottomrule
\end{tabular}
\end{subtable}
\end{table}

The construction in this appendix is the log-sum-exp merge of partial softmax outputs that underlies online softmax~\citep{milakov2018online}, memory-efficient and fused attention~\citep{rabe2021selfattention,dao2022flashattention}, blockwise and Ring Attention~\citep{liu2023blockwise,liu2024ringattention}, and split-K Flash-Decoding~\citep{dao2023flashdecoding}, applied to a cross-attention whose queries are replicated and whose keys and values are token-sharded.
We claim no methodological novelty for it; the appendix fixes notation, records the numerically stable reduction schedule and the backward pass that the implementation uses, and accounts for the communication of FLARE and FLARE++.

\subsection{Token-sharded representation}

Let the input to a token-mixing block be $X\in\mathbb{R}^{N\times C}$, with $C=HD$ the channel width for $H$ heads of dimension $D$; we omit the batch dimension, along which every operation below acts independently.
The context-parallel group contains $R$ ranks, and rank $r$ owns
\begin{equation}
    X_r\in\mathbb{R}^{N_r\times C},
    \qquad
    X=[X_0;\ldots;X_{R-1}],
    \qquad
    \sum_{r=0}^{R-1}N_r=N.
\end{equation}
Shards need not be spatially contiguous, since FLARE assigns no meaning to token order, and unequal $N_r$ are allowed as long as local masks exclude padding from the attention normalization.

All operations that act independently on tokens remain local: rank $r$ computes
\begin{equation}
    K_r=f_K(X_r),
    \qquad
    V_r=f_V(X_r),
    \qquad
    K_r,V_r\in\mathbb{R}^{H\times N_r\times D}.
\end{equation}
and the input and output projections, normalization, feed-forward networks, residual connections, and prediction head use the same sharding.
The learned query or seed tensors, of shape $\mathbb{R}^{H\times M\times D}$, are replicated across the group.

\subsection{Exact distributed FLARE encoder}

Consider a FLARE encoder with a replicated query tensor $Q\in\mathbb{R}^{H\times M\times D}$, stacking the per-head $Q_h$, and token-sharded keys and values.
On the concatenated token sequence, the desired result is
\begin{equation}
\label{eq:distributed_encoder_global}
    Z
    =\operatorname{softmax}\!\left(sQK^\top\right)V,
    \qquad
    K=[K_0;\ldots;K_{R-1}],
    \qquad
    V=[V_0;\ldots;V_{R-1}],
\end{equation}
where $s=D^{-1/2}$ and $Z\in\mathbb{R}^{H\times M\times D}$.
The global $K$ and $V$ are notation only and are never assembled.

Rank $r$ evaluates the encoder on its local token shard,
\begin{align}
    S_r &= sQK_r^\top
        \in\mathbb{R}^{H\times M\times N_r},\\
    L_r &= \log\sum_{j=1}^{N_r}\exp(S_{r,j})
        \in\mathbb{R}^{H\times M},
        \label{eq:local_lse}\\
    Z_r &= \operatorname{softmax}(S_r)V_r
        \in\mathbb{R}^{H\times M\times D}.
\end{align}
Because $Z_r$ is normalized only over shard $r$, the local outputs cannot be averaged uniformly.
With the global rowwise log-normalizer and the attention mass assigned to rank $r$,
\begin{equation}
\label{eq:global_lse}
    L=\log\sum_{r=0}^{R-1}\exp(L_r),
    \qquad
    \alpha_r=\exp(L_r-L),
\end{equation}
the exact global output is
\begin{equation}
\label{eq:distributed_encoder_merge}
    Z=\sum_{r=0}^{R-1}\alpha_r Z_r,
\end{equation}
since $\alpha_r$ restores the fraction of each globally normalized attention row that falls on shard $r$.

\subsection{Stable reductions and fused local attention}

Neither log-sum-exp above may be evaluated literally in finite precision, since both exponentiate unbounded arguments.
Both are instead evaluated through the exact shift identity
\begin{equation}
\label{eq:lse_shift}
    \log\sum_i\exp(x_i)
    =m+\log\sum_i\exp(x_i-m),
\end{equation}
with $m=\max_i x_i$, so that every exponential lies in $(0,1]$.
For the local reduction \eqref{eq:local_lse} the fused SDPA primitive already does this: its online softmax carries a running rowwise maximum and rescales its accumulators, so the $Z_r$ and $L_r$ it returns are computed in shifted form.
For the cross-rank reduction \eqref{eq:global_lse} we apply the shift with $m=L_{\max}$ in the following schedule.
An all-reduce maximum forms
\begin{equation}
    L_{\max}=\max_{r}L_r
    \in\mathbb{R}^{H\times M}
\end{equation}
at $\mathcal{O}(HM)$ values; each rank then computes, without communication,
\begin{equation}
    d_r=\exp(L_r-L_{\max}),
    \qquad
    U_r=d_r Z_r,
\end{equation}
with $d_r$ broadcast over the value dimension; two all-reduce sums form
\begin{equation}
    d=\sum_{r=0}^{R-1}d_r,
    \qquad
    U=\sum_{r=0}^{R-1}U_r,
\end{equation}
at $\mathcal{O}(HM)$ and $\mathcal{O}(HMD)$ values; and the exact output and log-normalizer follow elementwise on every rank,
\begin{equation}
\label{eq:stable_merge}
    Z=\frac{U}{d},
    \qquad
    L=L_{\max}+\log d.
\end{equation}
Since $d=\exp(-L_{\max})\sum_r\exp(L_r)$, $L_{\max}+\log d$ reproduces $L$ of \eqref{eq:global_lse} and $U/d$ reproduces \eqref{eq:distributed_encoder_merge}.
\autoref{tab:merge_forward_cost} lists the schedule with the cost of each step.
No step of the merge touches $N$, only the sum for $U$ carries the value dimension, and the merge arithmetic is elementwise on tensors the rank already holds, independent of the local token count.
FLARE performs one encoder invocation per mixer and FLARE++ two; neither all-gathers $X$, $K$, $V$, or $Y$, and the decoder adds no collective.
Aggregate network traffic and latency still depend on $R$ and on the collective implementation, which this per-rank accounting does not model.

A locally empty attention row must return $L_r=-\infty$ and $Z_r=0$ rather than NaN, which the shift maps to $d_r=U_r=0$; the merge is valid provided at least one shard holds a valid key for every row.
The score tensor $S_r$ is never materialized: a FlashAttention-backed SDPA primitive~\citep{dao2022flashattention,dao2024flashattention2} streams the local keys and values and returns both $Z_r$ and $L_r$.
Exposing $L_r$ is essential, because the output of an independently normalized local call alone cannot reconstruct the global result.

\subsection{FLARE and FLARE++ communication schedules}

In FLARE, the fixed queries $Q$ are replicated and one distributed encoder constructs the latent values,
\begin{equation}
    Z=\operatorname{Enc}_{\mathrm{dist}}(Q,\{K_r\},\{V_r\}),
\end{equation}
which the reductions in \eqref{eq:distributed_encoder_merge} leave replicated; rank $r$ then decodes locally,
\begin{equation}
\label{eq:local_flare_decode}
    Y_r=\operatorname{SDPA}(K_r,Q,Z),
\end{equation}
with no communication, because the query for each output token is local and both latent tensors are replicated.

FLARE++ invokes the same distributed encoder twice.
The first invocation constructs the synthesized routing queries,
\begin{equation}
    Q(X)
    =\operatorname{Enc}_{\mathrm{dist}}
      \left(\widetilde Q,\{\widetilde K_r\},\{\widetilde V_r\}\right),
\end{equation}
where the learned latent seeds $\widetilde Q$ are replicated and the projected input tensors $\widetilde K_r,\widetilde V_r$ remain sharded.
The second invocation gathers the physical values,
\begin{equation}
    Z
    =\operatorname{Enc}_{\mathrm{dist}}
      \left(Q(X),\{K_r\},\{V_r\}\right),
\end{equation}
and each rank applies the local decode
\begin{equation}
    Y_r=\operatorname{SDPA}(K_r,Q(X),Z).
\end{equation}
FLARE thus uses one communicating encoder and one local decoder, FLARE++ two communicating encoders and one local decoder; the output stays token-sharded and enters the next block without reconstructing an $N$-token tensor.

\subsection{Backward pass}

The distributed encoder composes a purely local attention on rank $r$, which maps $Q,K_r,V_r$ to the shard-normalized pair $(Z_r,L_r)$, with the global merge \eqref{eq:stable_merge}, which maps $\{(Z_r,L_r)\}_r$ to the replicated pair $(Z,L)$.
Reverse mode follows the same decomposition: given the cotangents of $Z$ and $L$, the merge produces the cotangents of each rank's $Z_r$ and $L_r$.
The local derivative must account for both outputs; an output-only SDPA backward does not expose this general interface.
Handing the cotangent of $Z$ to a local backward directly would differentiate $R$ independently normalized attentions instead of the globally normalized one.
We write $\bar A=\partial\mathcal{L}/\partial A$ for the cotangent of any forward tensor $A$ with respect to the scalar loss $\mathcal{L}$.

\paragraph{Incoming gradient of the merge.}
The all-reduces in \eqref{eq:stable_merge} leave every rank holding its own copy $Z^{[r]}=Z$, $L^{[r]}=L$, and rank $r$ consumes its copy in the local decode \eqref{eq:local_flare_decode}.
Replication is a fan-out in the computational graph, so its adjoint is a sum: writing $\bar Z^{(r)}=\partial\mathcal{L}/\partial Z^{[r]}$ and $\bar L^{(r)}=\partial\mathcal{L}/\partial L^{[r]}$ for the cotangents rank $r$ computes from its own copy,
\begin{equation}
\label{eq:merge_backward_input}
    \bar Z
    =\sum_{r=0}^{R-1}\frac{\partial\mathcal{L}}{\partial Z^{[r]}}
    =\sum_{r=0}^{R-1}\bar Z^{(r)},
    \qquad
    \bar L
    =\sum_{r=0}^{R-1}\bar L^{(r)}.
\end{equation}
Because every rank needs $\bar Z$ and $\bar L$ in full, the reduction is an all-reduce sum of $\mathcal{O}(HMD)$ and $\mathcal{O}(HM)$ values.
In FLARE and FLARE++ the decoder reads $Z$ alone, so $\bar L=0$ in practice and its all-reduce can be omitted; we carry it so that the merge is differentiable in both outputs and may be composed hierarchically.

\paragraph{Differentiating the merge.}
In its unshifted form the merge is
\begin{equation}
\label{eq:merge_unshifted}
    Z=\sum_{r=0}^{R-1}Z_r\exp(L_r-L),
    \qquad
    L=\log\sum_{r=0}^{R-1}\exp(L_r),
\end{equation}
a convex combination of the local outputs with weights $\exp(L_r-L)$, already computed in the forward pass as $d_r/d$.
Everything below is rowwise over $(h,m)$.
Rank $r$ enters \eqref{eq:merge_unshifted} only through $Z_r$ and $L_r$, with partial derivatives
\begin{align}
    \frac{\partial Z}{\partial Z_r}
        &= \exp(L_r-L)\,I_D,
        \label{eq:merge_jac_zz}\\
    \frac{\partial L}{\partial L_r}
        &= \exp(L_r-L),
        \label{eq:merge_jac_ll}\\
    \frac{\partial Z}{\partial L_r}
        &= \exp(L_r-L)\left(Z_r-Z\right),
        \label{eq:merge_jac_zl}
\end{align}
the third following from $\partial\exp(L_q-L)/\partial L_r=\exp(L_q-L)\left(\delta_{qr}-\exp(L_r-L)\right)$: raising $L_r$ slides the convex combination from $Z$ towards $Z_r$.
The chain rule gives the local cotangents
\begin{align}
    \bar Z_r &= \exp(L_r-L)\,\bar Z
        \in\mathbb{R}^{H\times M\times D},
        \label{eq:merge_backward_z}\\
    \bar L_r &= \underbrace{\exp(L_r-L)\,\bar L}_{\text{through }L}
        +\underbrace{\exp(L_r-L)\left\langle\bar Z,Z_r-Z\right\rangle_D}_{\text{through }Z}
        \in\mathbb{R}^{H\times M},
        \label{eq:merge_backward_l}
\end{align}
where $\langle A,B\rangle_D$ is the inner product over the value dimension, one scalar per row.
Shard $r$ receives the fraction $\exp(L_r-L)$ of the output cotangent that its attention mass earned in the forward pass, and $\bar L_r$ collects the two paths by which $L_r$ reaches the loss, through the merged normalizer and through the reweighting of shard $r$ against the others.
Neither expression communicates, since rank $r$ already holds $L_r,Z_r$ from the forward pass and $Z,L,\bar Z,\bar L$ from the reductions.

\paragraph{Local backward.}
Rank $r$ then runs the single-device backward with $(\bar Z_r,\bar L_r)$ as incoming gradients and the saved $Q,K_r,V_r,Z_r,L_r$.
Composing it with \plaineqref{eq:merge_backward_z} and \plaineqref{eq:merge_backward_l} yields a closed form in the globally normalized probability block $P_r=\exp(S_r-L)$, with $L$ broadcast along the local keys.
Writing $c=\left\langle\bar Z,Z\right\rangle_D$, the gradients are
\begin{align}
    \bar V_r &= P_r^\top\bar Z,\\
    \bar S_r &= P_r\odot
      \left(
        \bar Z V_r^\top-c+\bar L
      \right),\\
    \bar K_r &= s\,\bar S_r^\top Q,\\
    \bar Q &= s\sum_{r=0}^{R-1}\bar S_r K_r,
\end{align}
with $c$ and $\bar L$ broadcast over the $N_r$ local keys.
The composition collapses because $\exp(S_{r,j}-L_r)\exp(L_r-L)=\exp(S_{r,j}-L)$: the shard-normalized backward driven by $(\bar Z_r,\bar L_r)$ equals the globally normalized backward driven by $\bar Z$.
A conventional local backward on $\bar Z$ would instead use shard-normalized probabilities and $c=\left\langle\bar Z,Z_r\right\rangle_D$, dropping the between-shard term of \eqref{eq:merge_backward_l}.
For $\bar L=0$, an implementation can reuse a fused SDPA backward if it accepts the merged $Z$ and $L$ in place of the local pair, reconstructing $P_r$ without materializing the score matrix.
For nonzero $\bar L$, the additional $P_r\odot\bar L$ term must also be included; this requires the generalized derivative above.
Writing $\bar Q^{(r)}=s\,\bar S_r K_r$, the query gradient $\bar Q=\sum_r\bar Q^{(r)}$ is the only collective in this stage, one all-reduce sum of $\mathcal{O}(HMD)$ values, while $\bar K_r$ and $\bar V_r$ stay on the shard that owns them; gradients of replicated parameters are synchronized across the data- and context-parallel groups as usual.
\autoref{tab:merge_backward_cost} lists three backward collectives for the general two-output derivative, or two when the zero $\bar L$ reduction is omitted.
All act on $N$-independent latent quantities, with the only $N$-dependent work the fused local backward.

For FLARE++, each rank's decoder contributes to the cotangent of the replicated $Q(X)$; these contributions are summed across the group as in \eqref{eq:merge_backward_input} and added to the reduced $\bar Q$ from the physical-value encoder.
The resulting global query cotangent is passed once to the query-synthesis encoder, without reducing it a second time, and propagated by the same two-stage construction to local gradients for $\widetilde K_r$, $\widetilde V_r$ and a synchronized gradient for the seeds $\widetilde Q$.
Each encoder invocation must save its own output and log-normalizer.

\subsection{Measured scaling}
\label{app:multigpu_scaling}

\autoref{tab:cp_scaling} reports the measurements summarized in \autoref{sec:multigpu}, taken on a single node of four NVIDIA H100 GPUs; we did not have access to more ranks, so the table does not speak to cross-node interconnects.

\begin{table}[ht]
\centering
\caption{
Context-parallel strong scaling on NVIDIA H100 GPUs, FP16, at $C=128$, $H=8$, $B=8$.
Both columns are efficiencies normalized within a row against that row's single-rank measurement, so ideal scaling is $1.00$ in each: $E$ is the usual parallel efficiency in step time, and $E_{\mathrm{mem}}$ is the per-rank peak-memory reduction divided by the rank count.
$M=64$ and $M=128$ agree to within $1.2\%$ on efficiency and $0.1\%$ on memory, so only $M=128$ is shown.
This is not a comparison between FLARE and FLARE++.
}
\label{tab:cp_scaling}
\setlength{\tabcolsep}{6pt}
\input{figs/cp_scaling_table}
\end{table}

Efficiency is measured over $30$ timed steps per configuration and normalized within each model against its single-rank step time; one entry exceeds unity by about a percent.
The table reports timing estimates without uncertainty intervals.
Within these measurements, sharding reduces per-rank peak memory and step time close to proportionally to the rank count.
Memory is reported as a within-row ratio rather than in gigabytes because absolute peak memory is not comparable between this harness and the single-GPU harness of \autoref{fig:time_memory}, which is the sole source of absolute memory in this paper: the context-parallel harness reports a single-rank FLARE peak $1.89\times$ the single-GPU figure while agreeing on FLARE++ to $3\%$ and on step time to $5\%$, a discrepancy whose cause remains unresolved.
Within-row ratios describe scaling in the context-parallel harness, but do not establish why its absolute memory differs from the single-GPU harness.

\section{Dataset and protocol details}
\label{app:dataset_details}

\subsection{Benchmark datasets}
\label{app:benchmark_datasets}

\autoref{tab:dataset_tasks} summarizes each benchmark: the task, the discretization, the feature counts, and the splits.
The four standard benchmarks use the splits and resolutions of Transolver~\citep{wu2024transolver}, and DrivAerML-40K uses the split of the FLARE study~\citep{puri2026flare}; Appendix~D of that study describes each dataset and its source in full.
\autoref{fig:gallery_2d} shows one test case of each two-dimensional benchmark, and \autoref{fig:gallery_drivaerml} shows a full-surface DrivAerML geometry.

\begin{table}[ht]
\centering
\caption{
Benchmark tasks.
Point counts, feature counts, and splits of the first five rows are reproduced from Table~4 of the FLARE study~\citep{puri2026flare}; the last row is the full-surface task of \autoref{sec:drivaerml_scaling}.
}
\label{tab:dataset_tasks}
\setlength{\tabcolsep}{3pt}
\begin{tabular}{@{}>{\raggedright\arraybackslash}p{2.0cm} >{\raggedright\arraybackslash}p{4.6cm} >{\raggedright\arraybackslash}p{2.1cm} r c c@{}}
\toprule
\textbf{Benchmark} & \textbf{Task} & \textbf{Grid} & \textbf{Points} & \shortstack{\textbf{In/Out}\\\textbf{features}} & \shortstack{\textbf{Train/Test}\\\textbf{cases}} \\
\midrule
Elasticity & Stress field of an elastic solid from the point cloud of its geometry & 2D unstructured & 972 & 2 / 1 & 1000 / 200 \\
Darcy & Pressure of flow through a porous medium from its permeability field on an $85\times85$ grid & 2D structured & 7,225 & 1 / 1 & 1000 / 200 \\
Airfoil & Mach number field around deformed NACA-0012 profiles on a $221\times51$ body-fitted mesh & 2D structured & 11,271 & 2 / 1 & 1000 / 200 \\
Pipe & Horizontal velocity in pipes of varying centerline on a $129\times129$ mesh & 2D structured & 16,641 & 2 / 1 & 1000 / 200 \\
DrivAerML-40K & Surface pressure of morphed DrivAer vehicles from a fixed $40$K-cell subsample per geometry, used for both training and evaluation & 3D unstructured & 40,000 & 3 / 1 & 387 / 97 \\
DrivAerML (full surface) & Surface pressure and wall shear stress from cell coordinates and normals; $100$K random cells per training step, the full mesh at test & 3D unstructured & 8.2M & 6 / 4 & 400 / 50 \\
\bottomrule
\end{tabular}
\end{table}

\paragraph{The two DrivAerML tasks.}
\label{app:drivaerml_discussion}
The paper uses DrivAerML~\citep{ashton2024drivaerml} in two settings whose numbers are not comparable.
DrivAerML-40K, the last column of \autoref{tab:pde_benchmark_summary}, follows the FLARE study~\citep{puri2026flare}: $40$K surface cells are sampled once per geometry, the same fixed subsample serves for training and evaluation, the target is surface pressure, the split is $387/97$, and the metric is relative $L^2$ error.
The full-surface task of \autoref{sec:drivaerml_scaling} instead trains on a fresh random subsample of each geometry at every step and evaluates on the complete surface mesh.
The full-surface DrivAerML experiment uses the split of \citet{alkin2025abuptscalingneuralcfd}: $400$ training and $50$ test geometries, with the $34$ validation geometries unused.
Each surface cell carries its coordinates, scaled to the training-split bounding box, and its unit normal as input; the target is the four-channel vector of mean pressure and mean wall shear stress, standardized per channel by the training-split mean and standard deviation.
The training loss and the reported metric are the mean squared error of the standardized channels, averaged over cells and channels and multiplied by $100$; the test value is computed on the complete $8.2$M-cell surface of each test geometry and averaged over the test set.
Every model uses the backbone of \autoref{tab:config_shared} with $C=128$ and $H=8$, the same optimizer, peak learning rate, one-cycle schedule, and $500$ epochs as the standard benchmarks, FP16 execution with RMSNorm in place of LayerNorm, weight decay $10^{-4}$, and context parallelism over two H100 GPUs.
Each training step takes one geometry and a uniform random subset of $100$K of its cells drawn without replacement, redrawn at every step; no stratified or strided partitioning of the mesh is used.
Inference runs the full mesh through the model in a single forward pass under context parallelism, with no chunked decoding, for every model compared.

\subsection{Training and model configuration}
\label{app:standard_pde_configs}

\autoref{tab:config_shared} gives the settings held fixed across all datasets and mixers, and \autoref{tab:config_dataset} gives the per-dataset settings for the models we train under the matched backbone: full self-attention, Transolver, Transolver++, Transolver-3, Luna, AB-UPT, FLARE, and FLARE++.
AB-UPT uses $M=512$ anchor points on every dataset.
No hyperparameter is tuned per mixer.
Set Transformer uses the same settings as a complete architecture, and PerceiverIO, GNOT, and LNO are reproduced from the FLARE study~\citep{puri2026flare} under its configurations (\autoref{tab:comparison_models}).
All author-run accuracy results reported here use a single seed.
Experiments use NVIDIA H100 GPUs.
FP32 runs use LayerNorm, while mixed-precision runs use RMSNorm following the stability behavior observed in our runs~\citep{zhang2019root,micikevicius2018mixed}.
The same choice applies to the per-head normalizations inside FLARE++, which are $\mathcal N_a$ on the seeds $\widetilde Q_h$ and the routing keys $K_h$ and $\mathcal N_0$ on the shared $\widetilde V_h=\widetilde K_h$ and the fixed query $Q^{\mathrm f}_h$; FLARE++ accumulates the gated sum in FP32 and initializes every gate logit at $\gamma_h=0.25$, and the simplified variant of \autoref{app:simplified_flarepp} applies $\mathcal N_a$ to $\widetilde Q_h$ and $\widetilde K_h$ only (\autoref{tab:mixer_ops}).
Every feature projection is initialized near the identity, $W=I+\Delta W$ (\autoref{app:flarepp_identity_init}).

\begin{table}[ht]
\centering
\caption{
Settings held fixed across all datasets and all matched-backbone mixers.
The input and output projections follow the FLARE study~\citep{puri2026flare}.
}
\label{tab:config_shared}
\setlength{\tabcolsep}{6pt}
\begin{tabular}{ll}
\toprule
\textbf{Setting} & \textbf{Value} \\
\midrule
\multicolumn{2}{@{}l}{\emph{Architecture}} \\
Channel width $C$ & 128 \\
Attention heads $H$ & 8 \\
Head width $D=C/H$ & 16 \\
Input/output projection depth & 2 layers, following \citet{puri2026flare} \\
Output projection normalization & enabled \\
Block normalization & pre-norm LayerNorm \\
Block feed-forward network & GELU MLP \\
Feed-forward MLP ratio & 2.0 \\
\midrule
\multicolumn{2}{@{}l}{\emph{Optimization}} \\
Optimizer & AdamW, $\beta_1=0.9$, $\beta_2=0.999$, $\varepsilon=10^{-8}$ \\
Peak learning rate & $10^{-3}$ \\
Schedule & one-cycle \\
Exponential moving average & enabled \\
\midrule
\multicolumn{2}{@{}l}{\emph{Execution}} \\
Precision & FP32 \\
Hardware & NVIDIA H100 GPU \\
\bottomrule
\end{tabular}
\end{table}
\begin{table}[ht]
\centering
\caption{
Per-dataset settings.
The depth $B$ is chosen to suit the size of each benchmark and is shared by every mixer on that dataset, as is the latent budget $M$ (latent subsequence length, or slices for the Transolver family); AB-UPT uses $M=512$ anchor points throughout.
Full self-attention has no latent budget.
Batch size is the global batch; all runs here use a single device.
}
\label{tab:config_dataset}
\setlength{\tabcolsep}{6pt}
\begin{tabular}{lccccc}
\toprule
\textbf{Dataset} & \textbf{Blocks $B$} & \textbf{Latents $M$} & \textbf{Batch size} & \textbf{Weight decay} & \textbf{Epochs} \\
\midrule
Elasticity      & 8 & 64  & 2 & $10^{-5}$ & 500 \\
Darcy           & 8 & 128 & 2 & $10^{-5}$ & 500 \\
Airfoil         & 8 & 64  & 2 & $10^{-5}$ & 500 \\
Pipe            & 2 & 64  & 2 & $10^{-5}$ & 500 \\
DrivAerML-40K   & 4 & 64  & 1 & $10^{-2}$ & 500 \\
\bottomrule
\end{tabular}
\end{table}

\subsection{Comparison models}
\label{app:comparison_model_construction}

\autoref{tab:comparison_models} lists every model of \autoref{tab:pde_benchmark_summary}, how it is evaluated, its latent budget, and where it departs from \autoref{tab:config_shared} and \autoref{tab:config_dataset}; \autoref{tab:param_counts} gives the resulting parameter counts.

\begin{table}[ht]
\centering
\caption{
Comparison models.
Mixers are substituted into the shared residual backbone of \autoref{tab:config_shared}; the latent budget $M$ and the depth $B$ take the per-dataset values of \autoref{tab:config_dataset}, and deviations are listed in the last column.
Complete architectures replace the backbone.
}
\label{tab:comparison_models}
\setlength{\tabcolsep}{3pt}
\begin{tabular}{@{}>{\raggedright\arraybackslash}p{3.0cm} >{\raggedright\arraybackslash}p{3.3cm} >{\raggedright\arraybackslash}p{2.4cm} >{\raggedright\arraybackslash}p{4.6cm}@{}}
\toprule
\textbf{Model} & \textbf{Evaluated as} & \textbf{Latent budget} & \textbf{Deviations from Tables~\ref{tab:config_shared}--\ref{tab:config_dataset}} \\
\midrule
Full self-attention & Mixer & None & None; not run on Pipe or DrivAerML-40K owing to quadratic cost. \\
FLARE, FLARE++ & Mixer & $M$ latent queries & None. \\
Transolver, Transolver++, Transolver-3 & Mixer & $M$ slices & None; variants without convolution. \\
Luna & Mixer & $M$ latent tokens & None. \\
AB-UPT & Mixer & $512$ anchors & Latent budget fixed at $512$ anchors on every dataset; at $32$--$128$ anchors its loss is markedly worse. \\
Set Transformer & Complete architecture; an ISAB replaces each block & $M$ inducing points & None. \\
PerceiverIO & Complete architecture; reproduced from the FLARE study & $512$ latents & $C=128$, $H=8$, $B=8$ latent self-attention blocks, MLP ratio $4$~\citep{puri2026flare}. \\
LNO & Complete architecture; reproduced from the FLARE study & $256$ latent modes & $C=128$ ($192$ on Elasticity), $H=8$, $4$ latent self-attention blocks ($8$ on Airfoil and Pipe)~\citep{puri2026flare}. \\
GNOT & Complete architecture; reproduced from the FLARE study & None (linear attention with expert gating) & $C=128$, $H=8$, $B=8$, MLP ratio $2$, $3$ experts~\citep{puri2026flare}. \\
\bottomrule
\end{tabular}
\end{table}

\begin{table}[ht]
\centering
\caption{
Parameter counts, in millions, of the author-run models of \autoref{tab:pde_benchmark_summary} at the settings of \autoref{tab:config_shared} and \autoref{tab:config_dataset}, including biases and normalization parameters, computed from the reference implementation.
The models reproduced from the FLARE study~\citep{puri2026flare} use that study's configurations and counts: PerceiverIO $1.87$M, LNO $0.92$M, and GNOT $4.88$M.
}
\label{tab:param_counts}
\setlength{\tabcolsep}{6pt}
\begin{tabular}{lccccc}
\toprule
\textbf{Model} & \textbf{Elasticity} & \textbf{Darcy} & \textbf{Airfoil} & \textbf{Pipe} & \textbf{DrivAerML-40K} \\
\midrule
Full self-attention & 1.16 & 1.16 & 1.16 & 0.36 & 0.63 \\
Set Transformer     & 1.76 & 1.83 & 1.76 & 0.52 & 0.93 \\
Luna                & 1.57 & 1.57 & 1.57 & 0.47 & 0.84 \\
Transolver          & 1.04 & 1.05 & 1.04 & 0.34 & 0.57 \\
Transolver++        & 0.92 & 0.94 & 0.92 & 0.30 & 0.51 \\
Transolver-3        & 1.04 & 1.05 & 1.04 & 0.34 & 0.57 \\
AB-UPT              & 1.16 & 1.16 & 1.16 & 0.36 & 0.63 \\
FLARE               & 1.09 & 1.16 & 1.09 & 0.35 & 0.60 \\
FLARE++             & 1.29 & 1.42 & 1.29 & 0.40 & 0.70 \\
\bottomrule
\end{tabular}
\end{table}

We evaluate Set Transformer and PerceiverIO as complete architectures because their block structure differs from the shared token-mixer backbone.
Set Transformer's induced set attention block (ISAB) contains two cross-attention blocks, each with its own projections, residual paths, normalization, and feed-forward network.
PerceiverIO instead encodes into a latent sequence, processes it with latent self-attention, and decodes at the output.
A comparison using only their attention operations would evaluate a different architectural adaptation.
Here, Set Transformer replaces each attention-plus-feed-forward block with an ISAB, while PerceiverIO and LNO use the latent-processing depths of their reported configurations.

\paragraph{Reproducibility of FLARE.}
The FLARE entries in \autoref{tab:pde_benchmark_summary} differ from Table~1 of the FLARE study~\citep{puri2026flare} because the two tables answer different questions.
That table compares complete surrogate architectures at an approximately matched parameter count: FLARE uses residual MLPs for its key and value projections and its feed-forward block, and reduces the channel width to $C=64$ to offset their cost, on the hypothesis that mapping smooth fields to smooth fields benefits from deeper residual networks.
Our comparison instead holds the residual depth, channel width, and feed-forward block fixed across mixers, uses linear key and value projections inside every mixer, and lets the parameter count vary (\autoref{tab:param_counts}), so that the mixers are compared in isolation.
The FLARE study reports the same kind of identical-backbone comparison in its Appendix~G.1, and our FLARE results agree with those.

\paragraph{Transolver without convolution.}
Transolver supplements its physics-attention token mixer with convolution layers that aggregate over the structured grid~\citep{wu2024transolver}.
The FLARE study compares Transolver with and without these layers and reports a significant regression without them on Darcy, Airfoil, and Pipe~\citep{puri2026flare}.
To keep the comparison isolated to the token mixer, we likewise remove the convolution layers from Transolver, Transolver++, and Transolver-3 (\autoref{tab:comparison_models}).
Our Transolver results are in line with the Transolver results of the FLARE study under the same setting.

\paragraph{Reproducibility of Transolver++.}
The Transolver++ entries in \autoref{tab:pde_benchmark_summary} are our runs under the shared protocol, and they are less accurate than the results reported in its paper~\citep{luotransolver++}.
We report the values we obtained.
The difficulty of matching the published numbers with the official release and its stated hyperparameters is documented independently by the FLARE study~\citep{puri2026flare}, by AB-UPT~\citep{alkin2025abuptscalingneuralcfd}, and by the \href{https://docs.nvidia.com/physicsnemo/26.03/physicsnemo/examples/cfd/external_aerodynamics/transformer_models/README.html}{NVIDIA PhysicsNeMo documentation}~\citep{nvidia_physicsnemo_transolver_2026}, whose external-aerodynamics experiments report no gains from Transolver++.

\paragraph{Reproducibility of Transolver-3.}
We also attempted to include Transolver-3~\citep{zhou2026transolver3} in the full-surface DrivAerML comparison of \autoref{sec:drivaerml_scaling}.
The official Transolver-3 repository provides only an FP32 training recipe, and we know of no stable FP16 or BF16 recipe for it; every attempt to train it on DrivAerML in 16-bit precision diverged.
Running it in FP32 would break the matched-precision protocol of that experiment and does not fit the memory budget of the full-surface evaluation, so Transolver-3 is omitted from Figures~\ref{fig:drivaerml_gpu_hours} and~\ref{fig:drivaerml_scaling}.

\section{Extended results and ablations}
\label{app:extended_results}

\subsection{Random-seed variance}
\label{app:seed_variance}

\autoref{tab:seed_variance} repeats the FLARE and FLARE++ Elasticity runs of \autoref{tab:pde_benchmark_summary} with ten random seeds each.
The FLARE++ test error has mean $0.39\%$ and standard deviation $0.021\%$, a coefficient of variation of about $5\%$, and ranges from $0.37\%$ to $0.43\%$ across seeds.
The FLARE test error has mean $0.71\%$ and standard deviation $0.081\%$, a coefficient of variation of about $11\%$, and ranges from $0.58\%$ to $0.86\%$.
The two ranges do not overlap, so every FLARE++ seed is below every FLARE seed, and every FLARE++ seed also remains below the next-best efficient model on Elasticity in \autoref{tab:pde_benchmark_summary}, Set Transformer at $0.57\%$.
\autoref{tab:pde_benchmark_summary} reports these two means for Elasticity; its other entries are single-seed runs.

\begin{table}[H]
\centering
\caption{
Train and test relative $L^2$ error (in \%) of FLARE and FLARE++ on Elasticity across ten random seeds.
The standard deviation is the sample standard deviation.
}
\label{tab:seed_variance}
\setlength{\tabcolsep}{0pt}
\begin{tabular*}{\textwidth}{@{\extracolsep{\fill}}lcccccccccccc@{}}
\toprule
\textbf{Seed} & 0 & 1 & 2 & 3 & 4 & 5 & 6 & 7 & 8 & 9 & \textbf{Mean} & \textbf{Std} \\
\midrule
\multicolumn{13}{@{}l}{\textit{FLARE}} \\
Train & 0.383 & 0.460 & 0.521 & 0.466 & 0.491 & 0.528 & 0.664 & 0.449 & 0.576 & 0.496 & 0.503 & 0.077 \\
Test  & 0.577 & 0.683 & 0.728 & 0.663 & 0.711 & 0.760 & 0.865 & 0.636 & 0.790 & 0.701 & 0.711 & 0.081 \\
\midrule
\multicolumn{13}{@{}l}{\textit{FLARE++}} \\
Train & 0.235 & 0.239 & 0.231 & 0.234 & 0.259 & 0.250 & 0.252 & 0.222 & 0.236 & 0.248 & 0.241 & 0.011 \\
Test  & 0.368 & 0.395 & 0.378 & 0.383 & 0.431 & 0.408 & 0.394 & 0.369 & 0.369 & 0.411 & 0.391 & 0.021 \\
\bottomrule
\end{tabular*}
\end{table}

\subsection{Mixed-precision sensitivity}
\label{app:precision}

\begin{figure}[H]
\centering
\includegraphics[width=\linewidth]{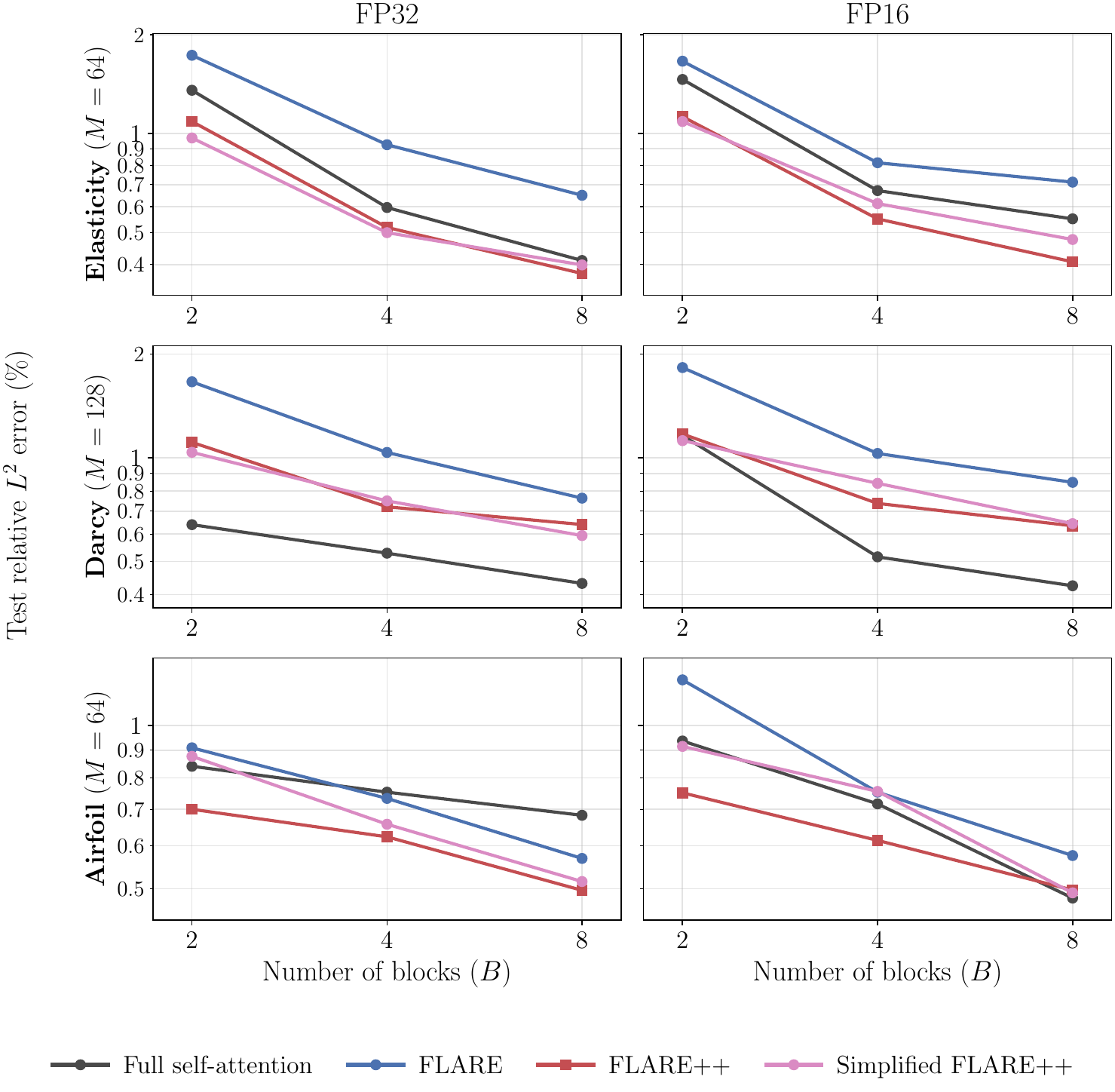}
\caption{
Test error against depth in FP32 (left) and FP16 (right), one row per benchmark and one line per token mixer, at each benchmark's latent budget from \autoref{tab:pde_benchmark_summary} ($M=128$ on Darcy, $M=64$ otherwise).
The two panels in a row share a vertical axis, so precision sensitivity appears as a change in the shape or ordering of the curves rather than their position; axes are not shared across rows.
}
\label{fig:precision}
\end{figure}

This section details the precision sweep summarized in \autoref{sec:experiments_precision}.
\autoref{fig:precision} repeats the depth sweep in both precisions for full self-attention, FLARE, FLARE++, and the simplified variant of \autoref{app:simplified_flarepp}, at the latent budget each dataset uses in \autoref{tab:pde_benchmark_summary}, on the three benchmarks with a full self-attention run in both precisions; every number is a single seed.
Pipe was also measured but is not plotted, since it has no FP32 full self-attention run.

The two execution settings change errors materially, while FLARE++ retains lower error than FLARE in every comparison.
The largest increase in this sweep is for full self-attention on Darcy at $B=2$, whose error rises by $82\%$; full self-attention on Airfoil at $B=8$ instead improves by $30\%$.
FLARE++ is the least precision-sensitive mixer on all three benchmarks: its FP16 error stays within $8\%$ of its FP32 error at every depth, against ratios of $0.88$--$1.33$ for FLARE, $0.95$--$1.23$ for the simplified variant, and $0.70$--$1.82$ for full self-attention.
The row that does move is the reference: in FP32, FLARE++ is more accurate than full self-attention on Airfoil at $B=8$ ($0.497$ against $0.683$), and in FP16 the order reverses ($0.497$ against $0.480$).
In principle a synthesized query is the more exposed construction, since a low-precision perturbation of the synthesis softmax moves the queries defining both rank-$M$ factors, whereas FLARE's queries are exact parameters.
The simplified variant shows this exposure: its FP16 error is $1.12$--$1.23\times$ its FP32 error on Elasticity, and on Pipe, which is not plotted, reaches $3.5\times$, $4.8\times$, and $12.9\times$ at $B=2,4,8$ while FLARE stays within $1.4\times$.
The fixed reference removes it on the three benchmarks plotted.
The settings change normalization as well as precision (\autoref{app:standard_pde_configs}), so this experiment measures their combined effect.

\clearpage

\subsection{Single-GPU time and memory}
\label{app:time_memory}

\autoref{fig:time_memory} reports forward-plus-backward time and peak memory of complete models at $B=8$, $C=128$, $H=8$ in FP16 on one NVIDIA H100, with the number of tokens varied from $1$K to $1$M.
Fused SDPA keeps every memory curve linear in $N$; full self-attention separates on time, and at $500$K tokens is two orders of magnitude slower than FLARE++.
FLARE++ costs $1.2$--$1.3\times$ FLARE's step time and $1.18\times$ its peak memory.
The time ratio is stable beyond $100$K tokens and is smaller than the $4/3$--$3/2$ increase in mixer arithmetic of \autoref{sec:flarepp} because the shared feed-forward network dilutes the mixer overhead.
Transolver-3 is run as a matched-backbone mixer without the tiling of \citet{zhou2026transolver3}; at $500$K tokens it needs $1.65\times$ FLARE's memory at $64$ slices and $2.47\times$ at $128$.
Absolute peak memory is comparable only within this harness; \autoref{app:multigpu_scaling} explains why the context-parallel measurements are reported as within-row ratios instead.

\subsection{DrivAerML depth, latent budget, and training cost}
\label{app:drivaerml_sweep}

\begin{figure}[!t]
\centering
\includegraphics[width=\linewidth]{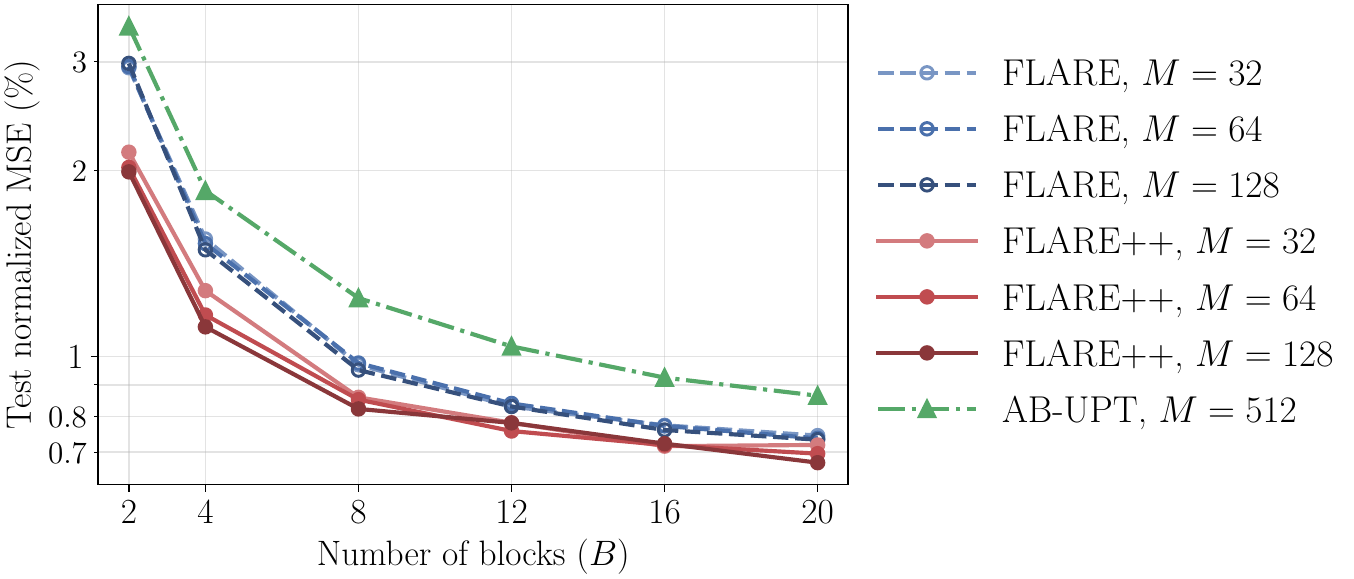}
\caption{
Test normalized MSE (in \%) against depth $B$ on DrivAerML's full $8.2$M-cell surface mesh, in FP16 under context parallelism, for FLARE, FLARE++, and AB-UPT.
The same runs are plotted against training cost in \autoref{fig:drivaerml_gpu_hours}.
}
\label{fig:drivaerml_scaling}
\end{figure}

\autoref{fig:drivaerml_scaling} plots the runs of \autoref{fig:drivaerml_gpu_hours} against depth, at every latent budget.
FLARE and FLARE++ each form a tight bundle across latent budgets, and FLARE++ lies below FLARE at every depth.
Both models continue improving with depth; with the best latent budget at each depth, FLARE's error falls by $49\%$ from two to four blocks, by a further $36\%$ from four to eight, and by $23\%$ more from eight to twenty.
Synthesized routing queries are therefore especially useful in shallow networks, while remaining beneficial as the backbone grows.

The GPU hours of \autoref{fig:drivaerml_gpu_hours} are estimated from separate single-GPU FP16 timing runs at the training configuration, as $500$ times the wall time of one steady-state epoch including data loading, and exclude compilation and full-mesh evaluation.
The ratio of $1.1$--$1.2\times$ between matched FLARE++ and FLARE runs is therefore an end-to-end figure, unlike the isolated step times of \autoref{app:time_memory}.
Below $2.5$ GPU hours, each FLARE run is the most accurate run at its cost, because the matching FLARE++ run costs $11$--$16\%$ more.

\subsection{Long Range Arena}
\label{app:lra}

FLARE++ uses the Transformer-block backbone, input and output projections, and feed-forward network of the FLARE LRA experiments.
The table combines baseline reruns from that study with published results; its caption identifies their provenance.
The comparison includes full self-attention~\citep{vaswani2017attention}; local attention and the original LRA baselines~\citep{taylong}; Reformer~\citep{kitaev2020reformer}; Sparse Transformer~\citep{child2019sparse}; Sinkhorn Transformer~\citep{tay2020sinkhorn}; Linformer~\citep{wang2020linformer}; Performer and FAVOR++~\citep{choromanski2020rethinking}; Funnel-Transformer~\citep{dai2020funnel}; Synthesizer~\citep{tay2021synthesizer}; linear attention~\citep{katharopoulos2020transformers}; Longformer~\citep{beltagy2020longformer}; BigBird~\citep{zaheer2020bigbird}; Norm attention~\citep{qin2022devil}; cosFormer~\citep{qin2022cosformer}; Nystr\"omformer~\citep{xiong2021nystromformer}; Skyformer~\citep{chen2021skyformer}; Hedgehog~\citep{zhang2024hedgehog}; Luna~\citep{ma2021luna}; and Transolver~\citep{wu2024transolver}.
FLARE uses the lightweight configuration of the FLARE study, with linear key/value projections, a GELU feed-forward network, and query/key normalization; FLARE++ uses the matched configuration with the additional query-synthesis branch.
The comparison covers attention-based encoders; it does not evaluate state-space sequence models.
As discussed by \citet{miralles2025lra}, LRA tasks also contain locality and positional biases, so we use this benchmark to assess transfer beyond PDEs.

\begin{table}[ht]
\centering
\normalsize
\setlength{\tabcolsep}{4.5pt}
\caption{
Accuracy (\%) on Long Range Arena (LRA) tasks \citep{taylong}.
Rows marked with $^\dagger$ use the reported result from \citet{zhang2024hedgehog}; the row marked with $^\ddagger$ uses the Luna result (projected length 256) reported by \citet{ma2021luna} under its own training setup; the other baseline results and the FLARE result reproduce the reruns reported alongside the FLARE arXiv paper~\citep{puri2026flare}.
The best result in each column is \textbf{bold} and the second best is \underline{underlined}.
}
\label{tab:experiments_lra}
\begin{tabular}{l|cccccc}
\toprule
Model & \textbf{ListOps} & \textbf{Text} & \textbf{Retrieval} & \textbf{Image} & \textbf{Pathfinder-32} & \textbf{Avg} \\
\midrule
Full self-attention
    & 36.70
    & 64.93
    & 77.18
    & 38.22
    & 70.52
    & 57.51 \\
\midrule
Local attention$^\dagger$
    & 15.82
    & 52.98
    & 53.39
    & 41.46
    & 66.63
    & 46.06 \\
Reformer$^\dagger$
    & 37.27
    & 56.10
    & 53.40
    & 38.07
    & 68.50
    & 50.67 \\
Sparse Transformer$^\dagger$
    & 17.07
    & 63.58
    & 59.59
    & 44.24
    & 71.71
    & 51.24 \\
Sinkhorn Transformer$^\dagger$
    & 33.67
    & 61.20
    & 53.83
    & 41.23
    & 67.45
    & 51.29 \\
Linformer (KV sharing)
    & 36.95
    & 51.74
    & 77.86
    & 42.82
    & 50.02
    & 51.88 \\
Performer
    & 35.90
    & 64.21
    & 68.42
    & 37.60
    & 53.83
    & 51.99 \\
Funnel-Transformer
    & \underline{38.50}
    & 61.17
    & 61.55
    & \textbf{53.10}
    & 49.98
    & 52.86 \\
Synthesizer$^\dagger$
    & 36.99
    & 61.68
    & 54.67
    & 41.61
    & 69.45
    & 52.88 \\
Linear attention
    & 17.95
    & \underline{66.00}
    & 71.84
    & 34.66
    & 75.00
    & 53.09 \\
Longformer$^\dagger$
    & 35.63
    & 62.85
    & 56.89
    & 42.22
    & 69.71
    & 53.46 \\
Linformer (headwise sharing)
    & 36.70
    & 53.00
    & 64.72
    & 43.42
    & 70.09
    & 53.59 \\
BigBird$^\dagger$
    & 36.05
    & 64.02
    & 59.29
    & 40.83
    & 74.87
    & 55.01 \\
Norm attention
    & 18.30
    & 63.08
    & 76.07
    & \underline{48.22}
    & 70.15
    & 55.16 \\
Performer (FAVOR++)
    & 36.00
    & 64.26
    & 76.74
    & 35.60
    & 69.67
    & 56.45 \\
cosFormer
    & 36.20
    & 64.59
    & 76.78
    & 41.52
    & 75.38
    & 58.89 \\
Nystr\"omformer$^\dagger$
    & 37.15
    & 65.52
    & 79.56
    & 41.58
    & 70.94
    & 58.95 \\
Skyformer$^\dagger$
    & \textbf{39.25}
    & 64.70
    & \underline{82.06}
    & 40.77
    & 70.73
    & 59.50 \\
Hedgehog$^\dagger$
    & 37.15
    & 64.60
    & \textbf{82.24}
    & 40.15
    & 74.16
    & 59.66 \\
Luna$^\ddagger$
    & 37.98
    & 65.78
    & 79.56
    & 47.86
    & \underline{78.55}
    & \textbf{61.95} \\
Transolver
    & 17.65
    & 60.95
    & 76.56
    & 34.50
    & 66.94
    & 51.32 \\
\midrule
\textbf{FLARE} \citep{puri2026flare}
    & 36.85
    & 65.23
    & 78.07
    & 36.86
    & 73.39
    & 58.08 \\
\textbf{FLARE++ (ours)}
    & 38.05
    & \textbf{66.56}
    & 79.37
    & 45.14
    & \textbf{78.86}
    & \underline{61.60} \\
\bottomrule
\end{tabular}
\end{table}

\clearpage
\section{Qualitative field predictions}
\label{app:qualitative}

\begin{figure}[H]
\centering
\includegraphics[width=\linewidth]{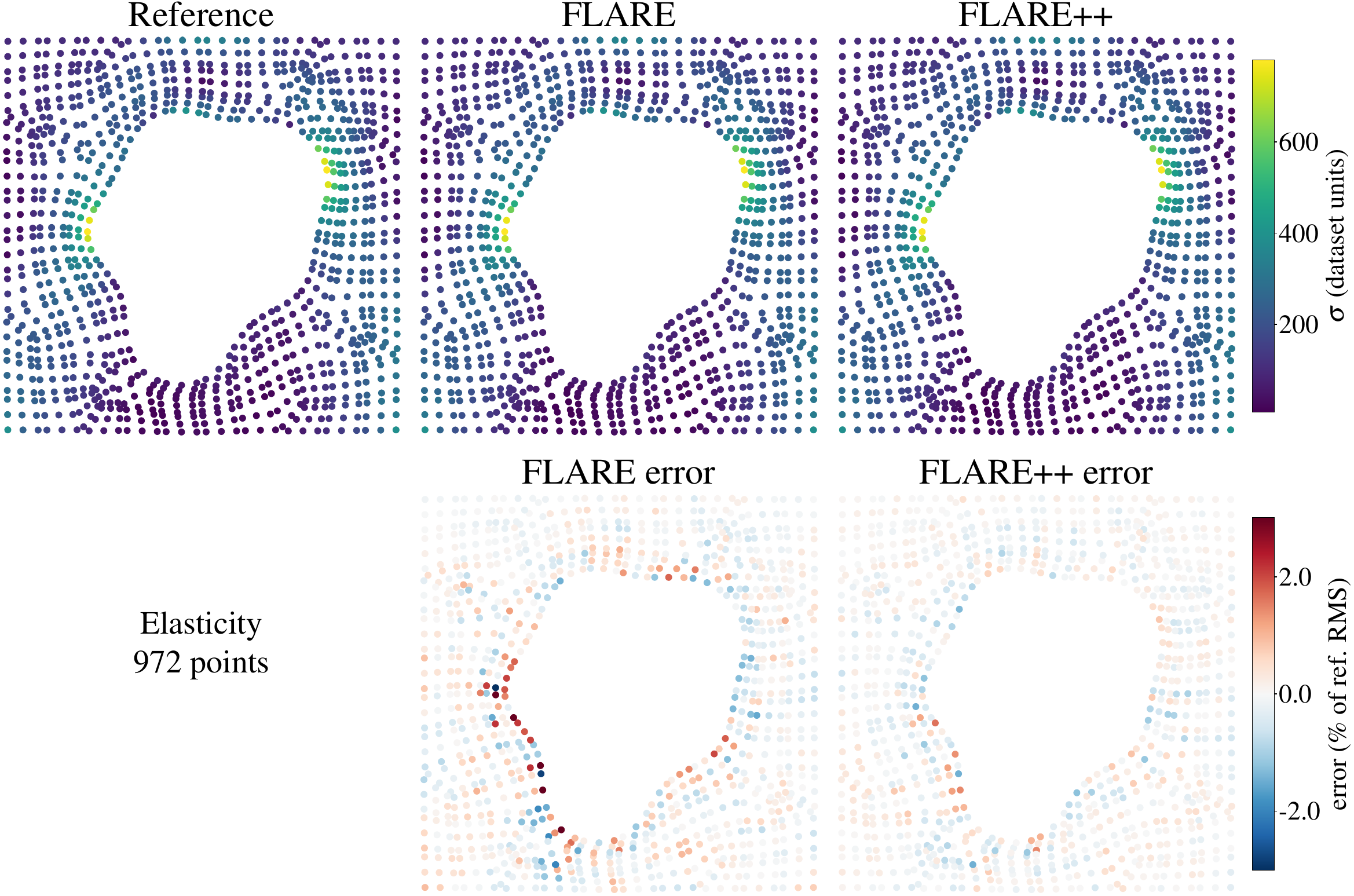}
\caption{
Elasticity: the von Mises stress $\sigma$ on the unstructured unit cell, in the units of the Geo-FNO release, which states none.
Top row, the reference solution and the two predictions; bottom row, each model's signed pointwise error in percent of the reference field's RMS.
}
\label{fig:vis_elasticity}
\end{figure}

\begin{figure}[H]
\centering
\includegraphics[width=\linewidth]{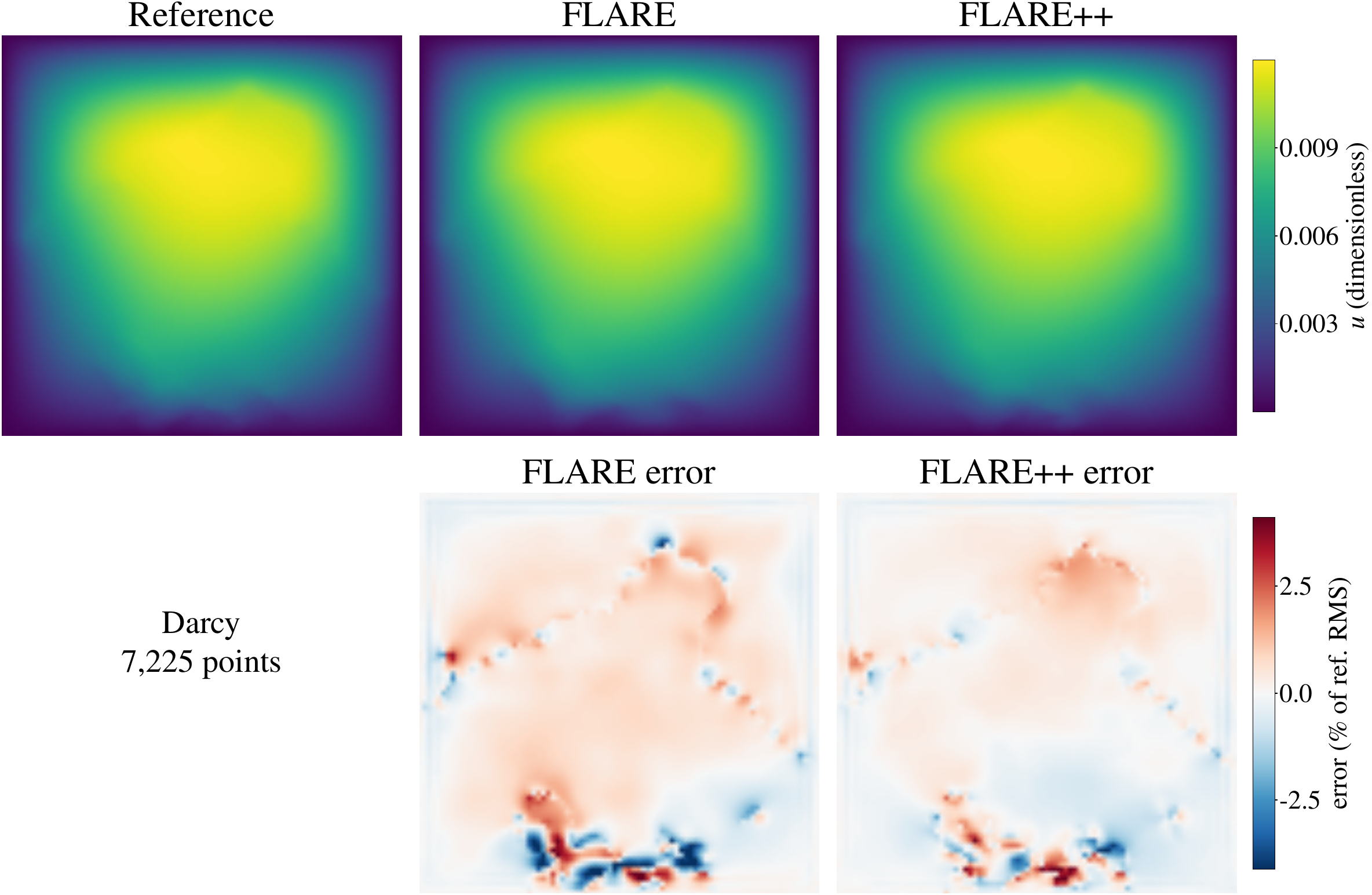}
\caption{
Darcy: the dimensionless pressure $u$.
Top row, the reference solution and the two predictions; bottom row, each model's signed pointwise error in percent of the reference field's RMS.
}
\label{fig:vis_darcy}
\end{figure}

\begin{figure}[H]
\centering
\includegraphics[width=\linewidth]{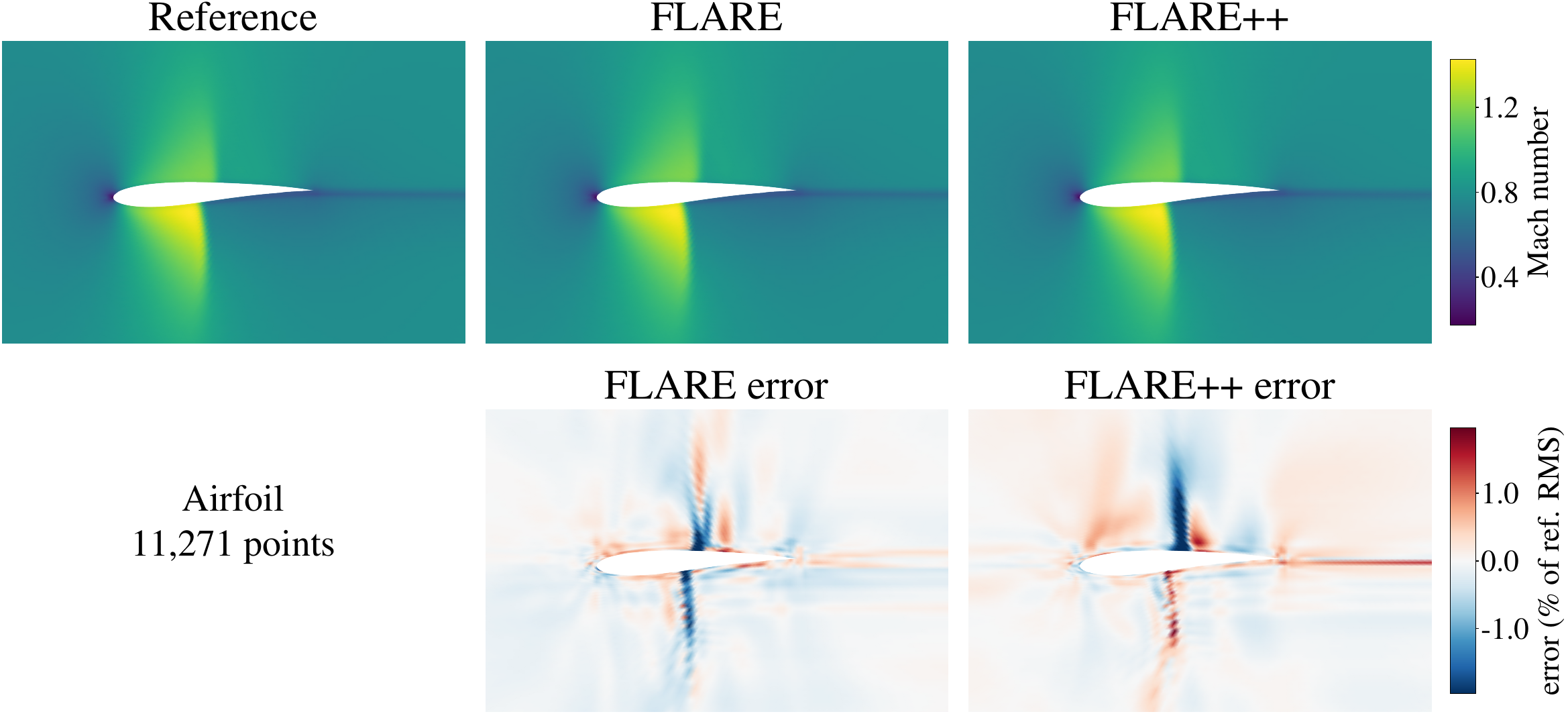}
\caption{
Airfoil: the Mach number (dimensionless) on the body-fitted mesh, cropped to the airfoil.
Top row, the reference solution and the two predictions; bottom row, each model's signed pointwise error in percent of the reference field's RMS over the full mesh.
}
\label{fig:vis_airfoil}
\end{figure}

\begin{figure}[H]
\centering
\includegraphics[width=\linewidth]{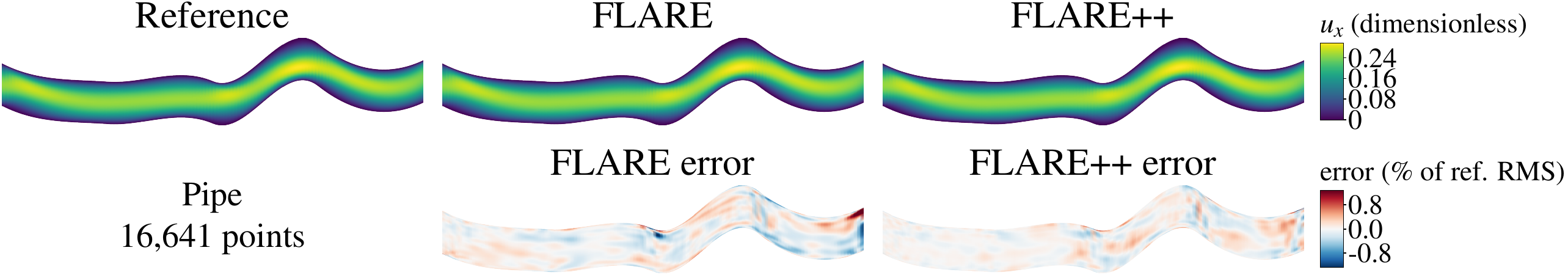}
\caption{
Pipe: the dimensionless streamwise velocity $u_x$.
Top row, the reference solution and the two predictions; bottom row, each model's signed pointwise error in percent of the reference field's RMS.
}
\label{fig:vis_pipe}
\end{figure}

\begin{figure}[H]
\centering
\includegraphics[width=\linewidth]{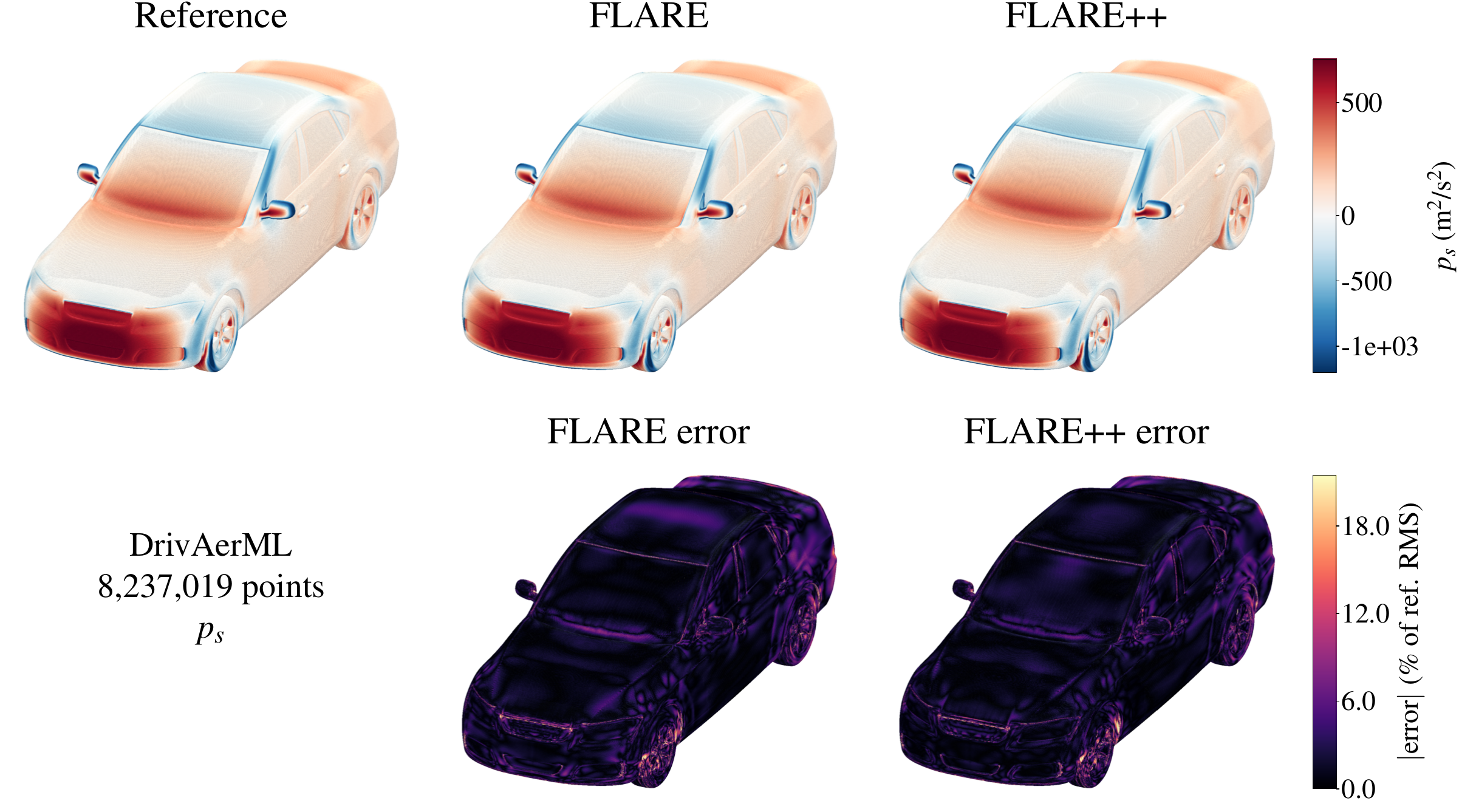}
\caption{
DrivAerML on the full surface mesh: the kinematic surface pressure $p_s=p/\rho$ in m$^2$/s$^2$.
Top row, the reference solution and the two predictions; bottom row, the magnitude of each model's pointwise error in percent of the reference field's RMS over the visible cells.
This is the full-surface task of \autoref{sec:drivaerml_scaling}, not DrivAerML-40K.
The case shown is the median over the first $10$ of the $50$ test geometries, the subset scored for this figure.
}
\label{fig:vis_drivaerml_ps}
\end{figure}

\begin{figure}[H]
\centering
\includegraphics[width=\linewidth]{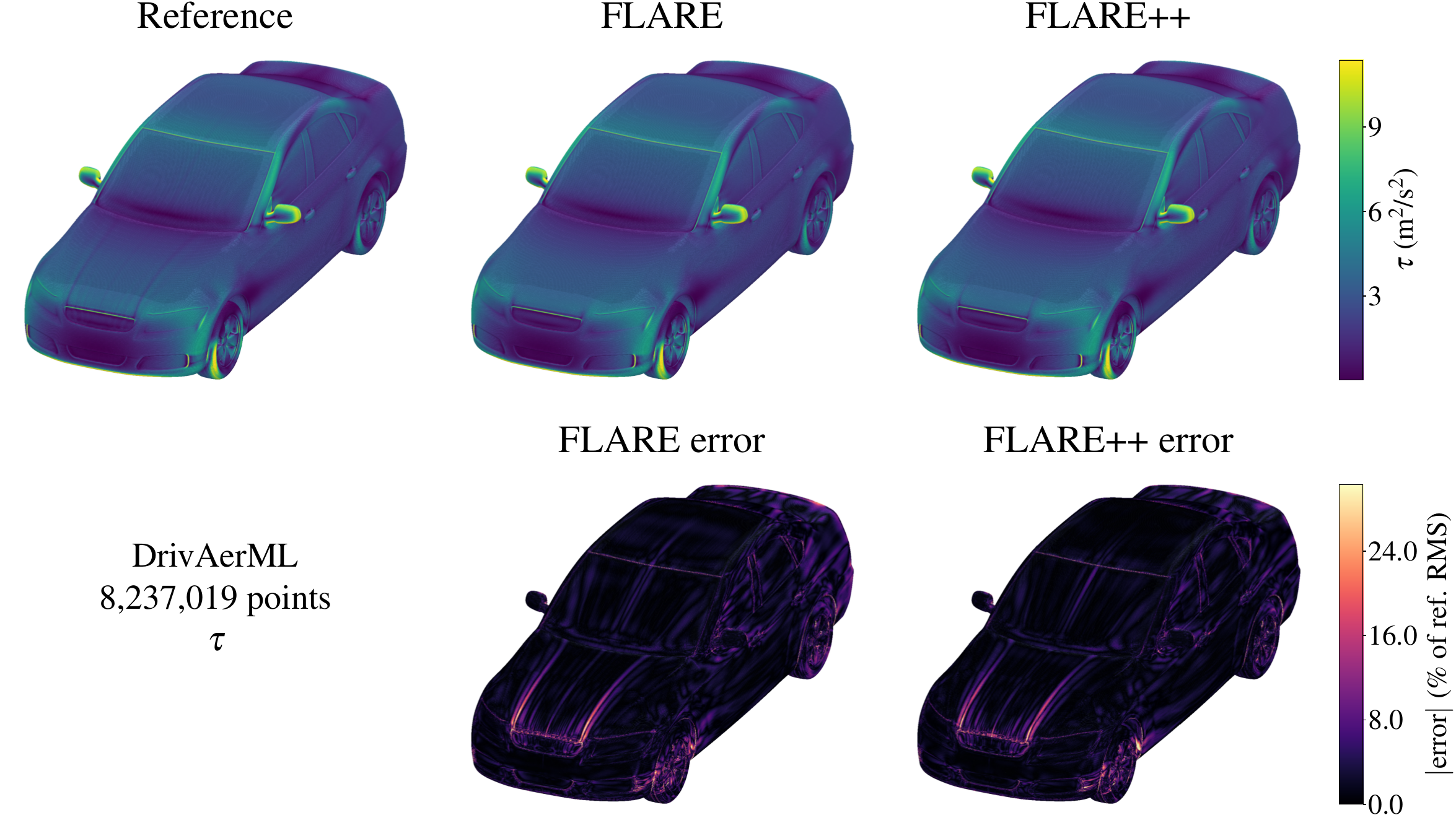}
\caption{
DrivAerML on the full surface mesh: the kinematic wall-shear magnitude $\tau=|\boldsymbol{\tau}_w|/\rho$ in m$^2$/s$^2$.
The case-selection subset is the same as in \autoref{fig:vis_drivaerml_ps}.
Top row, the reference solution and the two predictions; bottom row, the magnitude of each model's pointwise error in percent of the reference field's RMS over the visible cells.
}
\label{fig:vis_drivaerml_tau}
\end{figure}

Aggregate test errors do not reveal \emph{where} a surrogate fails.
Figures~\ref{fig:vis_elasticity}--\ref{fig:vis_drivaerml_tau} therefore show, for each benchmark, the reference solution and both predictions above the two pointwise error maps, on a test case selected by FLARE++'s median error, with both models on the same case.
The DrivAerML example is selected from ten scored test geometries, as specified in \autoref{fig:vis_drivaerml_ps}; it is not the median over the full test split.
The three field panels share one colour scale and the two error panels share another, so a smaller error appears as an emptier panel rather than a rescaled one.
The examples complement the aggregate errors with a view of their spatial distribution.
The error is unevenly distributed, concentrating near the void boundary on Elasticity, the permeability interfaces on Darcy, the leading edge and wake on Airfoil, the near-wall region on Pipe, and the edges and creases of the DrivAerML surface; the FLARE and FLARE++ maps on a given case look broadly similar and differ mainly in amplitude.

\paragraph{Rendering conventions.}
Every colour bar is labelled with the quantity it shows and its units.
Field panels show the predicted quantity as the dataset releases it: the Elasticity stress carries no units in the Geo-FNO release, the Darcy pressure and Pipe velocity are dimensionless, Airfoil shows the Mach number, and DrivAerML stores OpenFOAM's kinematic surface pressure $p_s=p/\rho$ and kinematic wall-shear magnitude $\tau=|\boldsymbol{\tau}_w|/\rho$, both in m$^2$/s$^2$.
Error panels show, at every point, the prediction minus the reference divided by the root-mean-square (RMS) of the reference field over the same case, in percent.
Two-dimensional panels show this signed value on a diverging scale centred at zero; surface panels show its magnitude on a sequential scale, with the RMS taken over the cells visible from the camera.
This is the pointwise analogue of the relative $L^2$ error in the tables: the RMS of an error panel over its case is that case's relative $L^2$ error in percent.
The two error panels of a figure share one colour scale whose limit is the $99.5$th percentile of the larger model's error magnitude.
Surfaces are rendered from cell centres and normals with VTK under fixed lighting and a parallel projection, with field limits at the $99.5$th percentile over visible points.
Grey marks regions with no data, including the Elasticity void.


\clearpage
\end{document}

%% file: vpack.tex
\usepackage{amsfonts,amssymb,amsbsy,amsmath}
\usepackage[inline]{enumitem}
\usepackage{physics}
\usepackage{cancel}        %
\usepackage{listings}      %

\usepackage{soul}          %
\usepackage{float}         %
\usepackage{graphicx}      %
\usepackage{caption}
\usepackage{subcaption}

\usepackage{xcolor}

\usepackage{hyperref} %

\usepackage{amsthm}

\renewcommand{\bar}[1]{\overline{#1}}

%% file: math_commands.tex
\usepackage{amsmath,amsfonts,bm}

\def\eqref#1{equation~\ref{#1}}

\def\plaineqref#1{\ref{#1}}

\def\1{\bm{1}}

\DeclareMathAlphabet{\mathsfit}{\encodingdefault}{\sfdefault}{m}{sl}
\SetMathAlphabet{\mathsfit}{bold}{\encodingdefault}{\sfdefault}{bx}{n}

%% file: figs/cp_scaling_table.tex
\begin{tabular}{llcccccc}
\toprule
\textbf{Model} & \textbf{$N$} & \multicolumn{2}{c}{$R=1$} & \multicolumn{2}{c}{$R=2$} & \multicolumn{2}{c}{$R=4$} \\
\cmidrule(lr){3-4} \cmidrule(lr){5-6} \cmidrule(lr){7-8}
 & & $E$ & $E_{\mathrm{mem}}$ & $E$ & $E_{\mathrm{mem}}$ & $E$ & $E_{\mathrm{mem}}$ \\
\midrule
\textbf{FLARE} & 500K & Ref. & Ref. & 0.98 & 0.97 & 0.94 & 0.97 \\
 & 1M & Ref. & Ref. & 1.01 & 0.98 & 0.98 & 0.97 \\
\textbf{FLARE++} & 500K & Ref. & Ref. & 0.97 & 0.96 & 0.92 & 0.95 \\
 & 1M & Ref. & Ref. & 0.96 & 0.97 & 0.97 & 0.96 \\
\bottomrule
\end{tabular}